\documentclass[11pt]{article}

\newif\ifanonymous
\ifdefined\submitmode\anonymoustrue\else\anonymousfalse\fi

\ifanonymous
  \usepackage[review,hyperref]{emnlp2026}
\else
  \usepackage[final,hyperref]{emnlp2026}
\fi
\usepackage{times}
\usepackage{microtype}
\usepackage{latexsym}
\usepackage{amsmath}
\usepackage{amssymb}
\usepackage{mathtools}
\usepackage{booktabs}
\usepackage{tabularx}
\usepackage{graphicx}
\graphicspath{{figures/}}
\usepackage{enumitem}
\usepackage{multirow}
\usepackage{xcolor}
\usepackage{colortbl}
\usepackage{tikz}
\usepackage{pgfplots}
\pgfplotsset{compat=1.18}
\definecolor{advint8}{RGB}{68,92,255}
\definecolor{advfp32}{RGB}{238,132,36}
\definecolor{advfp16}{RGB}{154,110,64}
\definecolor{advmlen}{RGB}{210,55,55}
\definecolor{advmllocal}{RGB}{130,28,130}
\usepackage{placeins}
\usepackage{balance}
\usepackage{float}
\usepackage{capt-of}
\usepackage{algorithm}
\usepackage{algorithmic}
\usetikzlibrary{arrows.meta,positioning,shapes.geometric,fit}

\ifanonymous
  \newcommand{\productname}{a widely-deployed AI code assistant}
  \newcommand{\idechat}{the IDE chat surface}
  \newcommand{\ideapp}{the IDE}
  \newcommand{\gateway}{the production inference gateway}
  \newcommand{\gatewayshort}{IG}
  
  \newcommand{\datasrc}{production}
  \newcommand{\Datasrc}{Production}
  \newcommand{\siblingsurfaces}{other IDE, CLI, web, mobile, and agent surfaces}
  \newcommand{\citeprodrouter}{\citep{prod-router-anon}}
\else
  \newcommand{\productname}{GitHub Copilot}
  \newcommand{\idechat}{VS Code Chat}
  \newcommand{\ideapp}{VS Code}
  \newcommand{\gateway}{GitHub Copilot API (CAPI)}
  \newcommand{\gatewayshort}{CAPI}
  
  \newcommand{\datasrc}{GitHub Copilot}
  \newcommand{\Datasrc}{GitHub Copilot}
  \newcommand{\siblingsurfaces}{the GitHub CLI, github.com, mobile (Android/iOS), and Coding Agent surfaces}
  \newcommand{\citeprodrouter}{\citep{copilot-capi}}
\fi

\title{HyDRA: Hybrid Dynamic Routing Architecture\\for Heterogeneous LLM Pools}

\author{
  \textbf{Aashna Garg} \quad
  \textbf{Siddharth Singha Roy} \quad
  \textbf{Jinu Jang}$^{*}$ \\
  \textbf{Federico Brancasi}$^{*}$ \quad
  \textbf{Giuseppe Cianci} \quad
  \textbf{Shengyu Fu} \\
  Microsoft \\
  \texttt{\{aashnagarg, ssingharoy, jinujang, fbrancasi, gcianci, shengyfu\}@microsoft.com} \\[2pt]
  {\footnotesize $^{*}$Equal contribution.}
}

\begin{document}
\maketitle

\begin{abstract}
Production LLM deployments maintain heterogeneous model pools spanning order-of-magnitude cost differences, yet existing routers make binary strong-vs-weak decisions coupled to specific model identities, requiring retraining on every catalog change. We present \textbf{HyDRA} (\textbf{Hy}brid \textbf{D}ynamic \textbf{R}outing \textbf{A}rchitecture), which predicts multi-dimensional capability requirements per query via a lightweight ModernBERT encoder with $K{=}4$ independent sigmoid heads (reasoning, code generation, debugging, tool use) and selects the cheapest model meeting those requirements through configuration-driven shortfall matching---fully decoupled from the model catalog, requiring zero retraining on catalog changes. On SWE-Bench Verified, \textbf{HyDRA} holds quality within 0.3 pp of the always-strong baseline at \textbf{54.1\% cost savings}---a \textbf{6$\times$} improvement over our prior production binary-v1 router---and generalizes across LiveCodeBench and BigCodeBench. On a held-out multilingual evaluation set, the same checkpoint retains \textbf{80.6\%} of oracle-routing quality at \textbf{37.5\% cost savings} (tunable to 56.1\% savings at 79.1\% retention), trading ${\sim}3$ quality points for large cost reductions relative to the strongest single model. A controlled A/B flight with close to 1M users per arm confirms improvements in latency, time-to-first-token, reliability, and engagement with no statistically significant user-visible degradation in measured metrics, alongside an estimated 7--20\% reduction in serving cost-of-goods-sold for the routed segment. \ifanonymous Deployed at production scale in the IDE chat surface of a widely-deployed AI code assistant,\else Deployed to all \productname{} \idechat{} users,\fi{} \textbf{HyDRA} is the first LLM-pool router to demonstrate cross-lingual routing consistency across 16 languages and 4 language groups.
\end{abstract}

\section{Introduction}

Production systems serving millions of users---code assistants, conversational agents, search assistants---now maintain pools of 10--15 LLMs, from lightweight models costing fractions of a cent per query to frontier reasoning models at 10--50$\times$ the price. The routing problem is straightforward: for each incoming query, select the cheapest model capable of producing a satisfactory response.

Despite its practical importance, routing remains surprisingly underexplored. The dominant deployed approach is infrastructure-level load balancing---entirely blind to what the user is asking~\citeprodrouter. Recent learned routers~\citep{ong2024routellm, ding2024hybrid, lu2024routing} improve on this but share three key limitations.

\ifanonymous
They are \textbf{model-coupled}---learning $f(\text{query})\to\text{model\_id}$ with model identities baked into training labels, so every catalog change forces retraining. They \textbf{collapse heterogeneous capability requirements onto a single axis}: binary routers~\citep{ong2024routellm} and scalar difficulty estimators~\citep{chen2023frugalgpt} cannot exploit a mid-tier model that is best-in-class on one dimension, a cost that grows with catalog heterogeneity (Table~\ref{tab:ablations}, Appendix~\ref{app:extended_eval}). And they \textbf{ignore language invariance}: published routers~\citep{ong2024routellm,ding2024hybrid,lu2024routing,zhang2025avengerspro} and concurrent systems~\citep{zhang2026mtrouter,zhang2026dialrouter,liu2026trouter,varshney2026llmrouter,madeyski2026triage} train and report only on English, while ``multilingual routing'' work targets token-level MoE experts~\citep{bandarkar2026moeroute} or programming languages~\citep{storek2025routesplain}.
\else
\textbf{Existing routers are model-coupled.} They learn $f(\text{query}) \to \text{model\_id}$ where model identities are embedded in training labels. When models are added, retired, or re-priced---a monthly occurrence---the router must be retrained.

\textbf{Existing routers collapse heterogeneous capability requirements onto a single axis.} A query requiring deep reasoning but trivial code output differs fundamentally from one needing sophisticated code generation but no reasoning, or one dominated by tool-use orchestration. Binary routers~\citep{ong2024routellm} and scalar difficulty estimators~\citep{chen2023frugalgpt} collapse this distinction onto a single strong-vs-weak score. On a single-task benchmark like SWE-Bench, where almost every query is reasoning- and code-heavy in similar proportions, the cost of this collapse is small (Table~\ref{tab:ablations}, Appendix~\ref{app:extended_eval}); the cost grows with workload and \emph{catalog} heterogeneity, where a scalar router cannot exploit a mid-tier model that is best-in-class on one dimension when it is added to the pool.

\textbf{Existing routers do not address language invariance.} Production coding assistants serve a global user base, yet published learned routers~\citep{ong2024routellm,ding2024hybrid,lu2024routing,zhang2025avengerspro} train and report exclusively on English benchmarks, and concurrent multi-turn~\citep{zhang2026mtrouter,zhang2026dialrouter} and pre-routing~\citep{liu2026trouter,varshney2026llmrouter,madeyski2026triage} systems do not evaluate cross-lingual behavior. Concurrent work using ``multilingual routing''~\citep{bandarkar2026moeroute} studies token-level expert routing inside one MoE model, and Routesplain~\citep{storek2025routesplain} treats ``multilingual'' as programming languages. We are not aware of a prior LLM-pool routing system that evaluates routing consistency across natural-language script families.
\fi

We propose \textbf{HyDRA}, built on three ideas: \textbf{multi-dimensional capability prediction} (a lightweight encoder predicts $K{=}4$ independent requirement scores per query---reasoning, code generation, debugging, tool use), \textbf{config-decoupled model matching} (model capabilities live in a YAML file, not learned weights; the router selects the cheapest model covering the predicted requirements via shortfall matching), and \textbf{language-invariant multilingual routing} (training on 16 languages across English, European, CJK, and other groups makes routing depend on task complexity, not language).

\textbf{HyDRA} is the second iteration of our deployed routing system, replacing an in-house binary strong-vs-weak ModernBERT classifier (\textbf{binary-v1}) that shipped in early 2026 and serves as our primary baseline. Our contributions are:
\begin{enumerate}[noitemsep,topsep=2pt]
  \item \textbf{Capability-decoupled routing via shortfall matching} (\S\ref{sec:arch}, \S\ref{sec:shortfall}). To our knowledge, the first router that fully decouples the learned predictor from the model catalog: predictions are over query \emph{requirements} along $K{=}4$ capability dimensions, and model selection is a configuration-driven shortfall-matching algorithm. Adding, removing, or repricing a model is a YAML edit---zero retraining, zero redeployment.
  \item \textbf{Multi-dimensional capability prediction with a structured labeling pipeline} (\S\ref{sec:labeling}). A single ModernBERT forward pass produces $K$ independent capability-requirement scores, trained on 50{,}159 dual-model LLM-judge labels over de-identified \datasrc{} telemetry, with position-swap debiasing.
  \item \textbf{Language-invariant routing across 16 languages and 4 language groups} (\S\ref{sec:exp_telemetry}). To our knowledge, the first \emph{LLM-pool} router with reported per-language quality and cost parity across English, European, CJK, and other language groups.
  \item \textbf{Production integration in \productname{}} (\S\ref{sec:deployment}): session-sticky routing, image hardgating, health-aware filtering, and zero-downtime model lifecycle management on \gateway{} infrastructure.
  \item \textbf{End-to-end empirical validation} (\S\ref{sec:experiments}): cost-quality Pareto sweeps on SWE-Bench Verified (matching the always-strong baseline within 0.3 pp at 54.1\% savings), cross-benchmark generalization to LiveCodeBench and BigCodeBench, and a controlled large-scale 50/50 A/B flight (\S\ref{sec:exp_abtest}).
\end{enumerate}

\section{Related Work}
\label{sec:related}

\ifanonymous
Prior LLM routers are predominantly binary (strong-vs-weak) and model-coupled: RouteLLM~\citep{ong2024routellm}, Hybrid LLM~\citep{ding2024hybrid}, and ZOOTER~\citep{lu2024routing} bind routing to training-time model identities, the last requiring $N$ forward passes. Cascades and ensembles (FrugalGPT~\citep{chen2023frugalgpt}, AutoMix~\citep{madaan2024automix}, EcoAssistant~\citep{zhang2024ecoassistant}, LLM-Blender~\citep{jiang2023llmblender}) add latency or $N\times$ cost, and mixture-of-experts~\citep{jiang2024mixtral} routes tokens within one model. Concurrent pre-routing work (MTRouter~\citep{zhang2026mtrouter}, DialRouter~\citep{zhang2026dialrouter}, TRouter~\citep{liu2026trouter}, LLM Router~\citep{varshney2026llmrouter}, Triage~\citep{madeyski2026triage}, RouteNLP~\citep{guo2026routenlp}) remains model-coupled and single-dimensional; R$^2$A~\citep{tang2026r2a} exposes an adversarial routing surface we discuss in \S\ref{sec:limitations}. A design-dimension comparison against ten systems is in Table~\ref{tab:competitive} (Appendix~\ref{app:competitive_design}); \textbf{HyDRA} is the only fully model-decoupled, multi-dimensional ($K{=}4$), multilingual (16 languages) entry, selecting a model in a single forward pass (${\sim}55$ms).
\else
\paragraph{LLM Routing.}
RouteLLM~\citep{ong2024routellm} trains classifiers on Chatbot Arena preference data for binary strong-vs-weak routing via matrix factorization that jointly embeds queries and models, coupling the router to training-time identities. Hybrid LLM~\citep{ding2024hybrid} uses a BERT difficulty predictor for binary routing. ZOOTER~\citep{lu2024routing} scores each candidate via a reward model, requiring $N$ forward passes. All are binary and model-coupled; \textbf{HyDRA} uses a single forward pass, predicts multi-dimensional \emph{requirements}, and stores model capabilities in configuration.

\paragraph{Cascading, ensembles, and MoE.}
FrugalGPT~\citep{chen2023frugalgpt}, AutoMix~\citep{madaan2024automix}, and EcoAssistant~\citep{zhang2024ecoassistant} cascade through models with verification, adding latency proportional to cascade depth and discarding partial generations. LLM-Blender~\citep{jiang2023llmblender} fuses responses from multiple models at $N\times$ cost; mixture-of-experts~\citep{jiang2024mixtral} routes tokens \emph{within} one model. \textbf{HyDRA} is pre-routing: the model is selected before generation, adding only encoder latency (${\sim}55$ms).

\paragraph{Concurrent routing work.}
MTRouter~\citep{zhang2026mtrouter} and DialRouter~\citep{zhang2026dialrouter} address multi-turn routing via learned trajectory outcome estimators and MCTS-derived policies---both model-coupled and trajectory-data-hungry. Pre-routing systems include TRouter~\citep{liu2026trouter} (query-conditioned latent task types), LLM Router~\citep{varshney2026llmrouter} (internal prefill activations as routing signals), Triage~\citep{madeyski2026triage} (code-health metrics for SWE tasks), and RouteNLP~\citep{guo2026routenlp} (closed-loop conformal cascading with distillation co-optimization); all remain model-coupled and single-dimensional. R$^2$A~\citep{tang2026r2a} shows adversarial suffix optimization can manipulate cost-aware routers, exposing a security surface we discuss in \S\ref{sec:limitations}. A design-dimension comparison against ten published routing systems is in Table~\ref{tab:competitive} (Appendix~\ref{app:competitive_design}); \textbf{HyDRA} is the only entry that is fully model-decoupled, multi-dimensional ($K{=}4$), and multilingual across 16 languages.
\fi

\section{Architecture}
\label{sec:arch}

\textbf{HyDRA} has three components: a capability requirement predictor (\S\ref{sec:predictor}), model capability profiles (\S\ref{sec:profiles}), and shortfall matching (\S\ref{sec:shortfall}). Figure~\ref{fig:architecture} illustrates the end-to-end flow.

\newcommand{\figArchitecture}{%
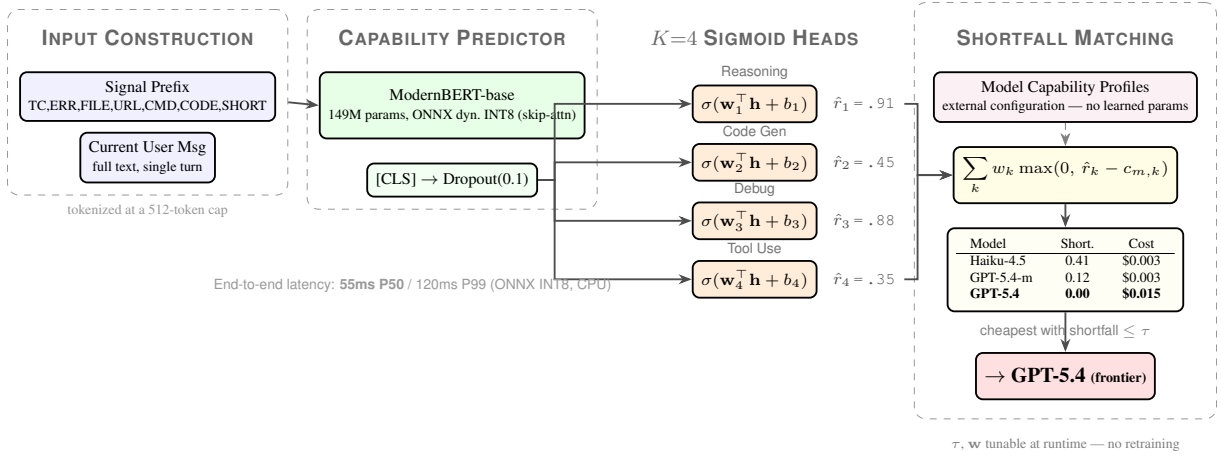
\begin{figure*}[t]
\centering
\ifanonymous\def\archwidth{0.75\textwidth}\else\def\archwidth{\textwidth}\fi
\resizebox{\archwidth}{!}{%
\begin{tikzpicture}[
    node distance=0.6cm and 0.9cm,
    box/.style={draw, rounded corners=3pt, minimum height=0.7cm, align=center, font=\scriptsize, thick},
    inputbox/.style={box, minimum width=1.5cm, fill=blue!6, font=\scriptsize},
    encoderbox/.style={box, minimum width=2.2cm, fill=green!10},
    headbox/.style={box, minimum width=1.4cm, minimum height=0.55cm, fill=orange!15, font=\scriptsize},
    scorebox/.style={minimum height=0.55cm, font=\scriptsize\ttfamily, text=black!70},
    matchbox/.style={box, minimum width=1.8cm, fill=yellow!12},
    profilebox/.style={box, minimum width=2.4cm, fill=purple!6, font=\scriptsize},
    outbox/.style={box, minimum width=1.8cm, fill=red!12, font=\small\bfseries},
    arr/.style={-{Stealth[length=2mm]}, thick, color=black!70},
    darr/.style={-{Stealth[length=2mm]}, densely dashed, color=black!50},
    groupbox/.style={draw, dashed, rounded corners=5pt, inner sep=5pt, color=black!40},
    lbl/.style={font=\scriptsize\sffamily\bfseries, color=black!60},
  ]

  \node[lbl] (stg1) {\footnotesize\textsc{Input Construction}};
  \node[inputbox, below=0.25cm of stg1] (i1) {Signal Prefix\\{\tiny TC,ERR,FILE,URL,CMD,CODE,SHORT}};
  \node[inputbox, below=0.15cm of i1] (i5) {Current User Msg\\{\tiny full text, single turn}};
  \node[groupbox, fit=(stg1)(i1)(i5), label={[font=\tiny,color=black!40]below:tokenized at a 512-token cap}] (inputgrp) {};

  \node[lbl, right=1.0cm of stg1] (stg2) {\footnotesize\textsc{Capability Predictor}};
  \node[encoderbox, below=0.25cm of stg2, minimum height=1.0cm] (enc) {ModernBERT-base\\{\tiny 149M params, ONNX dyn.\ INT8 (skip-attn)}};
  \node[box, below=0.35cm of enc, fill=green!5, minimum width=2.2cm, minimum height=0.5cm, font=\scriptsize] (cls) {[CLS] $\to$ Dropout(0.1)};
  \node[groupbox, fit=(stg2)(enc)(cls)] (encgrp) {};
  \draw[arr] (inputgrp.east) -- (enc.west);

  \node[lbl, right=1.0cm of stg2] (stg3) {\footnotesize\textsc{$K{=}4$ Sigmoid Heads}};
  \node[headbox, below=0.45cm of stg3] (h1) {$\sigma(\mathbf{w}_1^\top\mathbf{h}+b_1)$};
  \node[headbox, below=0.30cm of h1] (h2) {$\sigma(\mathbf{w}_2^\top\mathbf{h}+b_2)$};
  \node[headbox, below=0.30cm of h2] (h3) {$\sigma(\mathbf{w}_3^\top\mathbf{h}+b_3)$};
  \node[headbox, below=0.30cm of h3] (h4) {$\sigma(\mathbf{w}_4^\top\mathbf{h}+b_4)$};
  \node[scorebox, right=0.1cm of h1] (s1) {$\hat{r}_1$\,=\,.91};
  \node[scorebox, right=0.1cm of h2] (s2) {$\hat{r}_2$\,=\,.45};
  \node[scorebox, right=0.1cm of h3] (s3) {$\hat{r}_3$\,=\,.88};
  \node[scorebox, right=0.1cm of h4] (s4) {$\hat{r}_4$\,=\,.35};
  \node[font=\tiny\sffamily, color=black!55] at ([yshift=0.18cm]h1.north) {Reasoning};
  \node[font=\tiny\sffamily, color=black!55] at ([yshift=0.18cm]h2.north) {Code Gen};
  \node[font=\tiny\sffamily, color=black!55] at ([yshift=0.18cm]h3.north) {Debug};
  \node[font=\tiny\sffamily, color=black!55] at ([yshift=0.18cm]h4.north) {Tool Use};

  \draw[arr] (cls.east) -- ++(0.25,0) |- (h1.west);
  \draw[arr] (cls.east) -- ++(0.25,0) |- (h2.west);
  \draw[arr] (cls.east) -- ++(0.25,0) |- (h3.west);
  \draw[arr] (cls.east) -- ++(0.25,0) |- (h4.west);

  \node[lbl, right=1.2cm of stg3] (stg4) {\footnotesize\textsc{Shortfall Matching}};
  \node[profilebox, below=0.25cm of stg4] (prof) {%
    Model Capability Profiles\\
    {\tiny external configuration --- no learned params}};
  \node[matchbox, below=0.35cm of prof, minimum height=0.7cm] (sf) {%
    $\displaystyle\sum_k w_k \max(0,\, \hat{r}_k - c_{m,k})$};
  \node[box, below=0.35cm of sf, fill=yellow!5, minimum width=2.4cm, font=\tiny] (cand) {%
    \begin{tabular}{lcc}
    Model & Short. & Cost \\
    \hline
    Haiku-4.5 & 0.41 & \$0.003 \\
    GPT-5.4-m & 0.12 & \$0.003 \\
    \textbf{GPT-5.4} & \textbf{0.00} & \textbf{\$0.015} \\
    \end{tabular}};
  \node[font=\tiny\sffamily, below=0.05cm of cand, color=black!50] {cheapest with shortfall $\leq \tau$};

  \node[outbox, below=0.6cm of cand] (out) {$\to$ GPT-5.4 {\tiny (frontier)}};
  \draw[arr] (cand.south) -- (out.north);

  \node[groupbox, fit=(stg4)(prof)(sf)(cand)(out), inner sep=8pt] (matchgrp) {};

  \draw[arr] (s1.east) -- ++(0.2,0) |- (sf.west);
  \draw[arr] (s4.east) -- ++(0.2,0) |- (sf.west);
  \draw[darr] (prof) -- (sf);
  \draw[arr] (sf) -- (cand);

  \node[font=\tiny, color=black!50, below=0.15cm of matchgrp, align=center] {$\tau$, $\mathbf{w}$ tunable at runtime --- no retraining};
  \node[font=\tiny\sffamily, color=black!40, below=1.3cm of i5, xshift=4cm] {End-to-end latency: \textbf{55ms P50} / 120ms P99 (ONNX INT8, CPU)};
\end{tikzpicture}%
}
\caption{\textbf{HyDRA} architecture overview. \textbf{(1)~Input Construction}: a 7-flag signal prefix is concatenated with the current user message (512-token cap). \textbf{(2)~Capability Predictor}: ModernBERT-base produces a [CLS] embedding. \textbf{(3)~Sigmoid Heads}: $K{=}4$ independent heads predict per-dimension requirement scores $\hat{r}_k \in [0,1]$. \textbf{(4)~Shortfall Matching}: scores are compared against externally configured model profiles; the cheapest model with shortfall $\leq \tau$ is selected. Example scores shown are for the query ``Fix the race condition in WebSocket reconnect.''}
\label{fig:architecture}
\end{figure*}%
}
\figArchitecture

\subsection{Capability Requirement Predictor}
\label{sec:predictor}

Given query $q$, the predictor estimates $K$ scores $\hat{r}_1, \ldots, \hat{r}_K \in [0,1]$. We use ModernBERT-base~\citep{warner2024modernbert} (149M parameters). The [CLS] representation passes through dropout and $K$ independent linear heads:
\begin{equation}
  \hat{r}_k = \sigma\!\bigl(\mathbf{w}_k^\top \operatorname{dropout}(\mathbf{h}_{\mathrm{[CLS]}}) + b_k\bigr)
  \label{eq:pred}
\end{equation}
Total added parameters: $K \times 769 = 3{,}076$ for $K{=}4$---negligible relative to the encoder.

\paragraph{Input representation.}
The input concatenates a 7-flag \textbf{signal prefix} (turn-count bin, error/file/URL/command/code/short-message flags) with the \textbf{current user message}, tokenized at a 512-token cap. The predictor is single-turn: prior assistant responses, tool outputs, and repository state are excluded, keeping inference cheap and matching the information available before the LLM is called.

\paragraph{Training objective.}
Binary cross-entropy per dimension, with dimension-specific weights $\alpha_k$ (default 1.0):
\begin{equation}
\begin{split}
\mathcal{L} = -\frac{1}{\sum_k \alpha_k} \sum_{k} \alpha_k \big[\, & r_k \log \hat{r}_k \\
&+ (1{-}r_k)\log(1{-}\hat{r}_k) \,\big].
\end{split}
\end{equation}

\paragraph{Training recipe.}
\label{sec:multilingual_training}
Only one checkpoint is deployed---\textbf{HyDRA-Multi}, fine-tuned from \texttt{answerdotai/ModernBERT-base} on a single merged English+multilingual corpus: 5 epochs, batch size 32, lr $2{\times}10^{-5}$, cosine schedule with 10\% warmup, weight decay 0.01, fp16, seed 42; 6{,}270 steps in ${\sim}$42 min on one A100. A 2048-token variant ablation is in Table~\ref{tab:ctx_ablation} (Appendix~\ref{app:extended_eval}).

\subsection{Model Capability Profiles}
\label{sec:profiles}

Each model $m$ has $\mathbf{c}_m \in [0,1]^K$ and $\text{cost}_m$, stored in a YAML configuration file. Profiles are computed in two steps (Algorithm~\ref{alg:capability}, Appendix~\ref{app:algorithms}): a weighted average of public benchmark scores per dimension yields the raw per-model capability, then a pool-relative affine map rescales those raw scores into the requirement predictor's empirical score band measured on de-identified \datasrc{} data used for offline evaluation. Stored routing weights are compensated for differences in per-dimension band width so that operator intent is preserved.

\textbf{Step 1: Benchmark anchoring.}
For each dimension $k$, the raw capability is a weighted average of per-benchmark resolution rates, weighting each benchmark/subgroup by its importance $\alpha_b$ and the LLM-judge panel's per-dimension weight $\omega_{b,k}$ (full weights in Table~\ref{tab:capability_weights}, Appendix~\ref{app:benchmarks}). Each benchmark contributes only to the dimensions it can plausibly exercise---$\omega_{b,k}{=}0$ otherwise (e.g.\ $\tau^2$-bench~\citep{barres2025tau2bench} maps to reasoning and tool use; code benchmarks map to reasoning, code gen, and debugging; Table~\ref{tab:benchmark_mapping}).

\textbf{Step 2: Pool-relative normalization.}
Each model's raw score is affinely mapped into the requirement predictor's empirical score band $[\beta^{\text{lo}}_k, \beta^{\text{hi}}_k]$ (per-dimension percentiles on a held-out \datasrc{} set), pinning the weakest model to $\beta^{\text{lo}}_k$ and the strongest to $\beta^{\text{hi}}_k$:
\begin{equation}
  c_{m,k} = \beta^{\text{lo}}_k + \frac{\text{raw}_{m,k} - \min_j \text{raw}_{j,k}}{\max_j \text{raw}_{j,k} - \min_j \text{raw}_{j,k}} \cdot (\beta^{\text{hi}}_k - \beta^{\text{lo}}_k).
\end{equation}
Because band widths $\Delta_k$ differ across dimensions, operator-supplied weights $w_k$ are rescaled inversely to preserve stated intent:
\begin{equation}
  \tilde{w}_k = \frac{w_k / \Delta_k}{\sum_{k'} w_{k'} / \Delta_{k'}} \cdot \sum_{k'} w_{k'}.
  \label{eq:dimcomp}
\end{equation}

\subsection{Shortfall Matching}
\label{sec:shortfall}

The shortfall-matching algorithm (Algorithm~\ref{alg:routing}, Appendix~\ref{app:algorithms}) is the core routing decision procedure. Given predicted requirements $\hat{\mathbf{r}}$ and model profiles $\{(\mathbf{c}_m, \text{cost}_m)\}$:

\begin{equation}
  s_m \coloneqq \text{shortfall}(m) = \sum_{k=1}^{K} \tilde{w}_k \cdot \max(0, \hat{r}_k - c_{m,k})
  \label{eq:shortfall}
\end{equation}
where $\tilde{w}_k$ are band-compensated weights (Eq.~\ref{eq:dimcomp}). The $\max(0,\cdot)$ ensures surplus on one dimension does \emph{not} compensate for a deficit on another. The eligible set $\mathcal{E} = \{m : s_m \leq \tau\}$ is filtered by infrastructure health; the cheapest eligible model is selected, with fail-open to least-shortfall when $\mathcal{E}$ is empty. Both $\tau$ and $\mathbf{w}$ are runtime parameters---adjustable without retraining.

\subsection{Evaluation Metrics}
\label{sec:metrics}

\ifanonymous
We define three router-level metrics (pseudocode in Algorithm~\ref{alg:metrics}, Appendix~\ref{app:algorithms}). \textbf{Quality Retention (QR)}: resolution rate as a fraction of \emph{Oracle Routing}---the cheapest model resolving each query (Eq.~\ref{eq:qr}). \textbf{Cost Savings (CS)}: cost reduced vs.\ always routing to the most expensive model. \textbf{Misroute Rate (Mis)}: fraction of resolved queries a cheaper model would also have resolved (residual cost-saving opportunity), with cost ordering fixed per evaluation by unit prices. For the predictor we additionally report per-dimension MAE, RMSE, Pearson $r$, Spearman $\rho$, and binary accuracy at 0.5.
\else
We define three router-level metrics (pseudocode in Algorithm~\ref{alg:metrics}, Appendix~\ref{app:algorithms}). \textbf{Quality Retention (QR)}: resolution rate as a fraction of \emph{Oracle Routing}, the cheapest model that resolves each query (Eq.~\ref{eq:qr}). \textbf{Cost Savings (CS)}: fraction of cost reduced vs.\ always routing to the most expensive model. \textbf{Misroute Rate (Mis)}: fraction of queries where a cheaper model would also have resolved the query. The cost ordering is fixed once per evaluation by each model's input/output unit prices, rather than recomputed per query from realized token counts. Misroute therefore estimates residual cost-saving opportunity at unchanged observed resolution: these are cases where the router paid for more model than was necessary for a resolved outcome. For the predictor itself, we report per-dimension MAE, RMSE, Pearson $r$, Spearman $\rho$, and binary accuracy at threshold 0.5.
\fi
\begin{equation}
\text{QR} = \frac{\text{Res}_\text{router}}{\text{Res}_\text{oracle}} \times 100
\label{eq:qr}
\end{equation}

\section{Labeling Pipeline}
\label{sec:labeling}

\subsection{Context Tiering}

We route queries through three context tiers based on repository dependence: \textbf{T1} (${\sim}$4\%, explicit user-attached file references), \textbf{T2} (${\sim}$62\%, requires repo via tool calls or deictic references), and \textbf{T3} (${\sim}$34\%, self-contained). T3 queries proceed to dual-model generation; T1/T2 receive conservative synthetic defaults (constant requirement $0.8$ on every dimension), biasing toward stronger models. For public-repo T1/T2 queries we reconstruct context by cloning at the recorded commit, raising coverage to ${\sim}$45\%.

\subsection{Dual-Model Generation and Judging}

\ifanonymous
For each T3 query we issue two parallel generations---cheap \texttt{gpt-5.4-mini} vs.\ strong \texttt{gpt-5.3-codex}, conditioned on up to 10 prior turns. A \texttt{gpt-5.2-chat} judge scores both responses 1--5 on \{reasoning, code\_gen, debugging, tool\_use\} with a winner and quality-gap tag; it is run twice with positions swapped to cancel positional bias~\citep{zheng2023judging}, averaging per-dimension scores for the strong response.
\else
For each query (the last user turn of a sampled conversation), we issue two parallel generations conditioned on the system prompt and up to the prior 10 turns of context (assistant chunks truncated to 8{,}000 characters). The \textbf{cheap model} is \texttt{gpt-5.4-mini} (Chat Completions, max 4{,}096 tokens); the \textbf{strong model} is \texttt{gpt-5.3-codex} (Responses API, max 8{,}192 tokens); the \textbf{judge} is \texttt{gpt-5.2-chat} (JSON-mode Chat Completions). The judge sees the user query and both responses and scores each on a 1--5 scale across \{reasoning, code\_gen, debugging, tool\_use\} with a winner $\in \{A, B, \text{tie}\}$ and a quality\_gap tag. To cancel positional bias~\citep{zheng2023judging}, the judge is invoked twice with positions swapped, and per-dimension scores for the strong response are averaged across calls.
\fi

\subsection{Requirement Labels}

Let $\bar{s}^{\text{cheap}}_k$ and $\bar{s}^{\text{strong}}_k$ denote the position-debiased $1\text{--}5$ judge scores for the cheap and strong responses on dimension $k$. The requirement label measures where the strong model adds value:
\begin{equation}
  r_k = \max(0,\; \bar{s}^{\text{strong}}_k - \bar{s}^{\text{cheap}}_k) \,/\, 5 \;\; \in [0,0.8]
\end{equation}
If both score equally on a dimension, the requirement is zero, so the router learns to escalate precisely where the strong model adds value. (Per-model capability scores used in analysis normalize the same judge outputs: $c_k = (\bar{s}_k - 1)/4$.)

\subsection{Training Data Statistics}

\ifanonymous
\textbf{HyDRA-Multi} trains on a merged English+multilingual corpus of 50{,}159 labeled queries (40{,}128/5{,}016/5{,}015 train/val/test) drawn from de-identified \datasrc{} telemetry of users opted in to product-improvement data sharing (\S\ref{sec:ethics}), stratified across 16 languages---English plus 15 non-English buckets in the CJK / European / Other groups (per-group counts in Table~\ref{tab:lang_coverage}, Appendix~\ref{app:per_lang}). Mean requirement labels: reasoning $0.31$, code generation $0.22$, debugging $0.18$, tool use $0.14$. A separate multilingual human-labeled audit set (${\sim}$3.8K queries across 16 languages, 3 annotators each) provides judge-independent validation: human inter-annotator agreement is moderate-to-high (overall Krippendorff's $\alpha{=}0.64$), and the deployed predictor tracks the adjudicated human labels at least as well as a strong single-response LLM judge ($\alpha{=}0.40$ vs.\ $0.24$) at lower absolute error (MAE $0.15$ vs.\ $0.23$) (Appendix~\ref{app:human_eval}).
\else
\textbf{HyDRA-Multi} is trained on a single merged English+multilingual corpus with split sizes 40{,}128 train / 5{,}016 validation / 5{,}015 test (50{,}159 labeled queries total). Queries are drawn from de-identified \datasrc{} production telemetry for users who opted in to product-improvement data sharing (\S\ref{sec:ethics}). Non-English queries are assigned to a language bucket using script-range regexes on the user-message text, yielding ${\sim}$42K conversations / 89K turns across 15 non-English languages in three groups: CJK (zh, ja, ko; 250 convs.), European (fr, it, es, de, pl, pt, tr; 41.7K convs.), and Other (ru, ar, th, vi, id; ${\sim}$300 convs.); per-group counts are in Table~\ref{tab:lang_coverage} (Appendix~\ref{app:per_lang}). Sampling is stratified by language with a per-language cap of 5{,}000 conversations, taking the last turn per conversation with up to 10 prior turns as context. The full training corpus and held-out eval set (\S\ref{sec:exp_telemetry}) cover 16 languages. Label distribution on the labeled subset (mean $\pm$ std): reasoning $0.31 \pm 0.28$, code generation $0.22 \pm 0.24$, debugging $0.18 \pm 0.23$, tool use $0.14 \pm 0.20$.

We additionally curate a \textbf{multilingual human-labeled audit set} (${\sim}$3.8K queries across 16 languages, 3 annotators per query) as a judge-independent audit. Human inter-annotator agreement is moderate-to-high (overall Krippendorff's $\alpha{=}0.64$); the deployed predictor matches or exceeds a strong single-response LLM judge against the adjudicated human labels ($\alpha{=}0.40$ vs.\ $0.24$) while achieving lower absolute error (MAE $0.15$ vs.\ $0.23$). Composition, annotation protocol, and full agreement metrics are in Appendix~\ref{app:human_eval}.
\fi

\section{Evaluation}
\label{sec:evaluation}
\label{sec:experiments}

We evaluate \textbf{HyDRA} along four tracks: (1)~a held-out \textbf{multilingual evaluation set} of 8{,}040 \datasrc{} queries across 16 languages (\S\ref{sec:exp_telemetry}); (2)~\textbf{SWE-Bench Verified}~\citep{jimenez2024swebench} with real per-instance cost (\S\ref{sec:exp_swebench}); (3)~\textbf{cross-benchmark generalization} on LiveCodeBench~\citep{jain2024livecodebench} and BigCodeBench~\citep{zhuo2024bigcodebench} (\S\ref{sec:exp_crossbench}); and (4)~\textbf{competitive comparison} against RouteLLM~\citep{ong2024routellm}, Avengers Pro~\citep{zhang2025avengerspro}, Azure Foundry router~\citep{azurefoundry2026router}, and OpenRouter~\citep{openrouter2025} (\S\ref{sec:exp_competitive}); the A/B flight is in \S\ref{sec:prod_eval}. All offline numbers use the deployed \textbf{HyDRA-Multi} checkpoint (\S\ref{sec:multilingual_training}) at a 512-token cap (2048-token ablation in Appendix~\ref{app:extended_eval}); baselines (\emph{Always-Strong}, \emph{Always-Cheap}, \emph{binary-v1}, and \emph{Oracle}) and 8-model offline pool prices are in Table~\ref{tab:model_pool} (Appendix~\ref{app:extended_eval}).

\subsection{Multilingual Held-Out Evaluation}
\label{sec:exp_telemetry}
\label{sec:multilingual_results}

Table~\ref{tab:ml_routing} reports routing quality on the 8{,}040-query \datasrc{} eval set (5{,}007 English + 3{,}033 non-English, 16 languages) against all single-model baselines and the oracle on the four-model pool; per-language and per-group breakdowns are in Appendix~\ref{app:per_lang}.

\begin{table}[t]
\centering
\small
\begin{tabular}{lrr}
\toprule
\textbf{Router / Model} & \textbf{QR} & \textbf{CS} \\
\midrule
\rowcolor{blue!10}
\textit{Oracle Routing}              & \textit{100.0} & \textit{38.2} \\
\rowcolor{gray!15}
\textit{GPT-5.3-Codex (strong)}      & \textit{\phantom{0}83.4} & \textit{\phantom{0}9.9} \\
\textit{Sonnet 4.6 (cost anchor)}    & \textit{\phantom{0}78.5} & \textit{\phantom{0}0.0} \\
\rowcolor{green!12}
\textit{GPT-5.4-mini}                & \textit{\phantom{0}77.4} & \textit{70.3} \\
\textit{Haiku 4.5}                   & \textit{\phantom{0}75.0} & \textit{65.9} \\
\midrule
binary-v1                            & \phantom{0}77.5 & 58.1 \\
\textbf{HyDRA-Multi} ($\tau{=}0.05$) & \textbf{80.6} & \textbf{37.5} \\
\textbf{HyDRA-Multi} ($\tau{=}0.20$) & \textbf{79.1} & \textbf{56.1} \\
\bottomrule
\end{tabular}
\caption{Multilingual routing quality on the 8{,}040-query \datasrc{} eval set (5{,}007 English + 3{,}033 non-English; per-language split in Appendix~\ref{app:per_lang}). Four-model pool: Claude-Haiku-4.5, Claude-Sonnet-4.6, GPT-5.3-Codex, and GPT-5.4-mini. \textbf{QR} = quality retention vs.\ Oracle Routing (Eq.~\ref{eq:qr}). \textbf{CS} = cost savings vs.\ the costliest in-pool model (Claude-Sonnet-4.6). GPT-5.3-Codex is the strongest single model by quality; the two \textbf{HyDRA-Multi} rows report $\tau{=}0.05$ and $\tau{=}0.20$ operating points.}
\label{tab:ml_routing}
\end{table}

Quality retention by language group (Figure~\ref{fig:lang_invariance}, Appendix~\ref{app:per_lang}) stays tightly clustered across all four groups, ranging from $80.1\%$ to $80.9\%$ and remaining within $0.8$ points of the English baseline.

The strongest single model (GPT-5.3-Codex, 83.4 QR) is \emph{not} the costliest---it already saves 9.9\% over the Sonnet anchor---so why not route everything to it? At $\tau{=}0.05$, \textbf{HyDRA} gives up 2.8 QR points but nearly quadruples cost savings (37.5\% vs.\ 9.9\%), and the A/B flight (\S\ref{sec:exp_abtest}) shows this frontier choice also improves latency, reliability, and engagement---outcomes a fixed single-model policy cannot tune.

\subsection{SWE-Bench Verified}
\label{sec:exp_swebench}

\newcommand{\tabMain}{%
\begin{table}[t]
\centering
\small
\begin{tabular}{lrrrr}
\toprule
\textbf{Router} & \textbf{Res\%} & \textbf{QR} & \textbf{CS} & \textbf{Mis.} \\
\midrule
\rowcolor{blue!10}
\textit{Oracle Routing}    & \textit{86.2} & \textit{100.0} & \textit{---} & \textit{\phantom{0}0.0} \\
\rowcolor{green!12}
\textit{GPT-5.4-mini (cheap)}      & \textit{69.4} & \textit{80.5} & \textit{78.6} & \textit{\phantom{0}0.0} \\
Claude Haiku 4.5  & 69.4 & 80.5 & 11.7 & 69.4 \\
GPT-5.3 Codex     & 73.4 & 85.2 & 57.3 & 78.0 \\
GPT-5.4           & 73.4 & 85.2 & 25.7 & 85.0 \\
\rowcolor{gray!15}
\textit{Claude Sonnet 4.6 (strong)} & \textit{74.2} & \textit{86.1} & \textit{\phantom{0}0.0} & \textit{82.4} \\
\midrule
binary-v1         & 73.8 & 85.6 & \phantom{0}9.1 & 75.0 \\
\textbf{HyDRA} (peak, $\tau{=}0.01$) & \textbf{75.4} & \textbf{87.5} & \textbf{12.9} & \textbf{81.0} \\
\textbf{HyDRA} (cons., $\tau{=}0.24$) & \textbf{74.0} & \textbf{85.8} & \textbf{54.1} & \textbf{70.2} \\
\textbf{HyDRA} (agg., $\tau{=}0.64$)  & \textbf{71.0} & \textbf{82.4} & \textbf{72.5} & \textbf{19.8} \\
\bottomrule
\end{tabular}
\caption{SWE-Bench Verified (500 instances; 5-model pool). \textbf{QR} = quality retention vs.\ Oracle Routing (Eq.~\ref{eq:qr}; Oracle resolves 431/500). \textbf{CS} = cost savings vs.\ always-strong (Claude Sonnet 4.6). \textbf{Mis.} = misroute (\%). binary-v1 is restricted to (Sonnet, GPT-5.4-mini). Three \textbf{HyDRA} operating points illustrate $\tau$-tunability: (peak) exceeds Sonnet by 1.4 QR at 12.9\% savings; (cons.) stays within 0.3 pp of Sonnet at 54.1\% savings ($6\times$ binary-v1); (agg.) trades 5.1 QR for 72.5\% savings. Full sweep in Fig.~\ref{fig:pareto}.}
\label{tab:main}
\end{table}%
}
\tabMain

A 2-model decomposition of the 400 solvable instances (Figure~\ref{fig:swe_venn}, Appendix~\ref{app:extended_eval}) confirms a balanced strong/weak split at the balanced-allocation operating point ($\tau{=}0.175$). Per-instance significance tests for all operating points (Wilson intervals, paired-bootstrap cost intervals, McNemar tests against always-strong) are in Appendix~\ref{app:significance}; no \textbf{HyDRA} operating point differs significantly in quality from always-strong.

\subsection{Cross-Benchmark Generalization}
\label{sec:exp_crossbench}

The deployed predictor is trained only on labeled \datasrc{} queries (\S\ref{sec:labeling}) and never sees SWE-Bench Verified, LiveCodeBench, or BigCodeBench at training time. We therefore test whether the four capability dimensions, the model profiles, and the shortfall policy transfer to these off-distribution coding benchmarks using the same production router config, with no per-benchmark tuning. Per-router quality and cost savings are reported in Appendix~\ref{app:generalization_coding_benchmarks} (supporting per-model resolution and cost in Tables~\ref{tab:cross_benchmark} and~\ref{tab:token_savings}).

\subsection{Competitive Comparison}
\label{sec:exp_competitive}

We compare \textbf{HyDRA} on SWE-Bench Verified against research routers (RouteLLM-MF, RouteLLM-BERT, Avengers Pro) and commercial offerings (Azure Foundry router, OpenRouter \texttt{auto}). The headline comparison is \emph{matched-pool}: every router selects only from a common 3-model pool (GPT-5, GPT-5-mini, GPT-5.2), so QR and CS are directly comparable (Table~\ref{tab:competitive_matched}, Figure~\ref{fig:pareto_compare}). Avengers Pro's matched-pool settings are fit on held-out coding benchmarks (LiveCodeBench for aggressive, BigCodeBench for conservative)---a favorable condition for it, as those benchmarks are closer to SWE-Bench than production chat. RouteLLM supports only a binary model list, so its strong/weak threshold sweep is reported separately (Table~\ref{tab:routellm_comparison}); design-level dimensions are summarized in Table~\ref{tab:competitive}.

\newcommand{\tabCompMatched}{%
\begin{table}[t]
\centering
\small
\setlength{\tabcolsep}{4pt}
\begin{tabular}{lccc}
\toprule
\textbf{System} & \textbf{QR (\%)} & \textbf{CS (\%)} & \textbf{Mis. (\%)} \\
\midrule
\rowcolor{blue!10}
\textit{Oracle Routing}                                & \textit{100.0} & \textit{+68.9} & \textit{0.0} \\
\rowcolor{green!12}
\textit{GPT-5-mini (always-cheap)}                      & \textit{73.2} & \textit{$+$90.1} & \textit{$\phantom{0}$0.0} \\
Avengers Pro (Cons.)                          & 91.8 & $+$15.6 & 58.8 \\
OpenRouter Auto (c/q$\,{=}\,0$)               & 90.3 & $+$\phantom{0}4.9 & 68.4 \\
Azure Foundry (Quality)                       & 82.9 & $+$36.3 & 38.8 \\
OpenRouter Auto (c/q$\,{=}\,1$)               & 82.1 & $+$44.2 & 37.6 \\
Avengers Pro (Aggr.)                          & 81.6 & $+$46.7 & 35.6 \\
Azure Foundry (Balanced)                      & 76.0 & $+$66.2 & 17.6 \\
\rowcolor{gray!15}
\textit{GPT-5.2 (always-strong)}                        & \textit{93.4}           & \textit{$\phantom{+}$0.0}   & \textit{72.4} \\
\midrule
\textbf{HyDRA (cons.)} $\tau{=}0.703$         & \textbf{90.3} & $\boldsymbol{+}\textbf{16.2}$ & \textbf{57.8} \\
\textbf{HyDRA (agg.)}  $\tau{=}0.888$         & \textbf{84.2} & $\boldsymbol{+}\textbf{44.1}$ & \textbf{35.6} \\
\bottomrule
\end{tabular}
\caption{\textbf{Matched-pool comparison on SWE-Bench Verified} ($n{=}500$, pool: GPT-5, GPT-5-mini, GPT-5.2). All routers select only from this fixed pool, so QR / CS / Mis.\ are directly comparable across rows. Reference rows highlighted: \emph{Oracle} (blue) is the per-query upper bound, \emph{always-cheap} (green) is the GPT-5-mini floor, \emph{always-strong} (gray) is the GPT-5.2 ceiling that CS is measured against. \textbf{QR}: quality retention vs.\ Oracle Routing. \textbf{CS}: cost savings vs.\ always-strong (GPT-5.2). \textbf{Mis.}: misroute rate. \emph{Conservative} settings preserve quality at the cost of smaller savings; \emph{aggressive} settings push savings further at the cost of some quality. \emph{OpenRouter Auto} is shown at two points of its server-side cost/quality (c/q) tradeoff knob (integer scale, $0=$ max quality).}
\label{tab:competitive_matched}
\end{table}%
}
\ifanonymous\else\tabCompMatched\fi

\newcommand{\figParetoCompare}{%
\begin{figure*}[t]
\centering
\includegraphics[width=\textwidth]{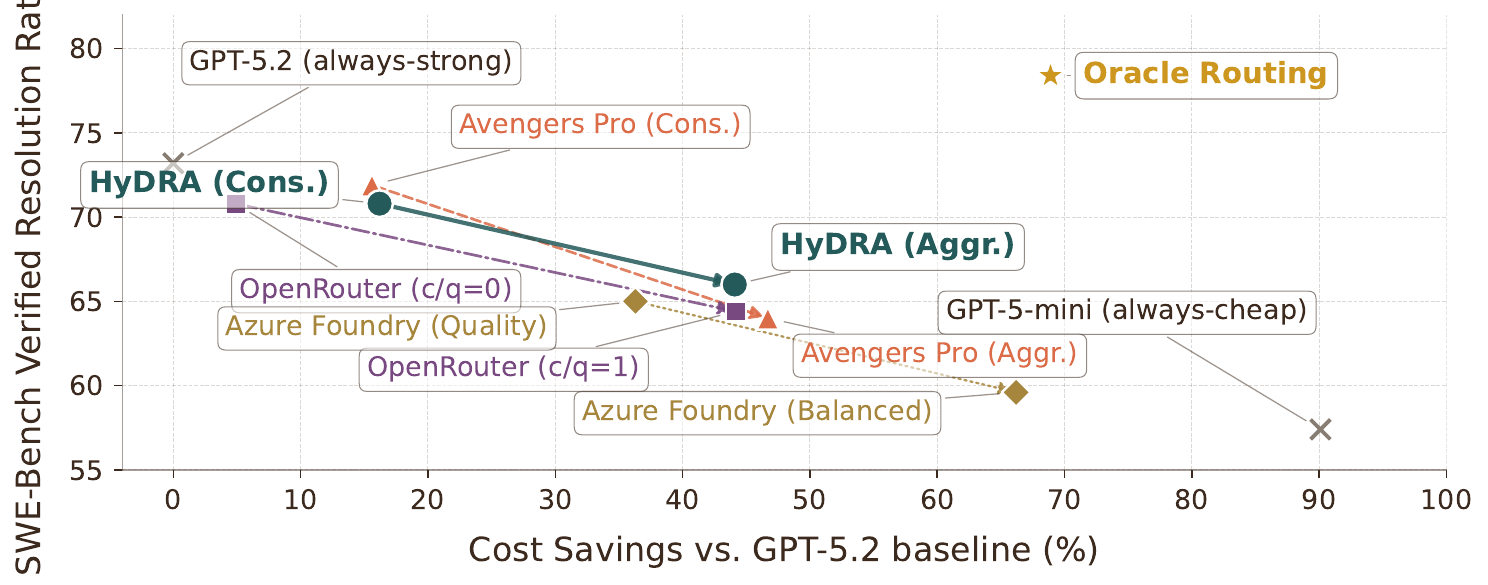}
\caption{\textbf{Pareto comparison on the matched 3-model pool} (GPT-5 / GPT-5-mini / GPT-5.2; data from Table~\ref{tab:competitive_matched}). \textbf{HyDRA} (cons.) ties OpenRouter Auto (cost/quality $=0$) on resolution rate (70.8\%) at $3.3\times$ the cost savings (16.2\% vs.\ 4.9\%); \textbf{HyDRA} (agg.) improves on Avengers Pro--Aggressive ($+2.0$ pp resolution rate for $-2.6$ pp CS), dominates OpenRouter Auto (cost/quality $=1$) ($+1.6$ pp resolution rate at equal CS), and outperforms both Azure Foundry operating modes on the reported QR/CS frontier. Avengers Pro--Conservative has the highest non-oracle resolution rate (71.9\%; $+1.2$ pp over \textbf{HyDRA} cons.) at near-equal CS, while \textbf{HyDRA}'s advantage is a competitive frontier from a model-decoupled, multi-dimensional router that does not require retraining on catalog changes. Upper-right is better.}
\label{fig:pareto_compare}
\end{figure*}%
}
\ifanonymous\else\figParetoCompare\fi

\ifanonymous
\noindent\textbf{Matched-pool result.} No router uniformly dominates the 3-model pool; the point is that one \emph{model-decoupled}, multi-dimensional router that absorbs catalog changes \emph{without retraining} stays on the competitive frontier against routers retrained or fit per pool. \textbf{HyDRA} (cons.) ties OpenRouter Auto (c/q$=0$) on QR (90.3) at $3.3\times$ the cost savings ($+$16.2 vs.\ $+$4.9); \textbf{HyDRA} (agg.) beats Avengers Pro--Aggressive ($+$2.6 pp QR for $-$2.6 pp CS), dominates OpenRouter Auto (c/q$=1$), and beats both Azure Foundry modes on the QR/CS frontier. Avengers Pro--Conservative attains the highest non-oracle QR (91.8) at near-equal CS, but its settings are fit on held-out coding benchmarks closer to SWE-Bench than production chat. Pairwise McNemar tests (Appendix~\ref{app:significance}) confirm \textbf{HyDRA} is statistically tied with the strongest competitor at each operating point.
\else
\noindent\textbf{Matched-pool result.} On the common 3-model pool, no router uniformly dominates. Avengers Pro--Conservative attains the highest non-oracle QR (91.8) at near-equal CS to \textbf{HyDRA} (cons.), while \textbf{HyDRA} (cons.) ties OpenRouter Auto (cost/quality $=0$) on QR (90.3) with substantially higher CS (+16.2 vs.\ +4.9). At the aggressive operating point, \textbf{HyDRA} improves over Avengers Pro--Aggressive in QR (+2.6 pp) while giving up 2.6 pp CS, dominates OpenRouter Auto (cost/quality $=1$) ($+2.1$ pp QR at equal CS), and outperforms both Azure Foundry operating modes on the reported QR/CS frontier. Thus the matched-pool result is best read as a competitive Pareto comparison: \textbf{HyDRA}'s empirical points are competitive with router-specific baselines, while its deployment advantage is that the same model-decoupled, multi-dimensional router supports arbitrary catalog changes without retraining. Pairwise McNemar tests (Appendix~\ref{app:significance}) confirm \textbf{HyDRA} is statistically tied with the strongest competitor at each operating point.
\fi

\subsection{Model Catalog Portability}
\label{sec:portability}
Table~\ref{tab:portability} (Appendix~\ref{app:competitive_design}) validates the decoupling claim on two catalog events: removing Claude Haiku 4.5, \textbf{HyDRA} reroutes via a YAML edit ($-$2.4 QR as Haiku's traffic falls back); adding a new mid-tier model, \textbf{HyDRA} routes to it on day one ($+$0.4 QR). binary-v1's QR is unchanged in both cases---it cannot perceive catalog changes without retraining.

\subsection{Latency}
\label{sec:latency}
Offline routing adds 55\,ms P50 / 120\,ms P99 on CPU (ModernBERT INT8 inference dominates; formatting and shortfall computation are sub-ms), well under 1\% of typical LLM generation time.

\section{Production Evaluation}
\label{sec:prod_eval}

The held-out \datasrc{} result that selects the deployed checkpoint is Table~\ref{tab:ml_routing} (\S\ref{sec:exp_telemetry}): \textbf{HyDRA-Multi} improves quality over binary-v1 (80.6 vs.\ 77.5 QR) while exposing a tunable cost--quality frontier on the same four-model pool. The offline tracks in \S\ref{sec:experiments} test whether that deployment decision generalizes.

\subsection{Production A/B Flight}
\label{sec:exp_abtest}

\ifanonymous
We run a controlled A/B flight comparing the deployed \textbf{HyDRA} arm against the prior production auto-mode control (binary-v1 plus the preceding heuristic policy) on the full auto-mode population of \idechat{}---close to 1M users per arm over a 14-day window, stratified by SKU. \textbf{HyDRA} delivers statistically significant improvements ($p<0.01$, Welch's $t$-test; delta-method variance for ratios) across latency, reliability, engagement, and efficiency with no statistically significant user-visible degradation in measured metrics: median time-to-complete drops $6.4\%$, time-to-first-token $4.6\%$, and the per-turn error rate $17.7\%$, while user-initiated turns rise $1.7\%$ and per-inference-request cost falls $2.3\%$. Versus serving the segment entirely with the flagship model, \textbf{HyDRA} yields an estimated $7$--$20\%$ COGS reduction; versus the prior control, aggregate segment COGS is roughly flat as higher request volume offsets the lower per-request cost. Multi-day engagement is flat ($|\Delta|\leq0.10\%$, $p>0.45$).
\else
We run a controlled A/B flight comparing the deployed \textbf{HyDRA} arm against the prior production auto-mode control---an in-house binary-v1 router combined with the heuristic-based auto-mode policy that preceded any learned routing---on the full auto-mode population of \idechat{}, with close to 1M users per arm over a 14-day window, stratified by SKU. The two-arm design controls for traffic-mix and seasonality effects single-arm before/after comparisons cannot. \textbf{HyDRA} delivers statistically significant improvements ($p<0.01$)\footnote{Significance is computed via a two-sample Welch's $t$-test on user-level metric aggregates; ratio metrics (per-user-per-day cost, per-turn latency, per-turn error rate) use delta-method variance to account for ratio-of-sums noise. User-level aggregation controls for within-user correlation across requests.} across latency, reliability, engagement, and efficiency with no statistically significant user-visible degradation in measured metrics. Relative to control, median time-to-complete drops $6.4\%$, time-to-first-token drops $4.6\%$, and the per-turn error rate drops $17.7\%$, while user-initiated turns rise $1.7\%$ and per-inference-request cost falls $2.3\%$. Relative to serving the auto-mode segment entirely with the flagship model, routing through \textbf{HyDRA} yields an estimated $7$--$20\%$ reduction in cost-of-goods-sold (COGS); against the prior router control, aggregate segment COGS is roughly flat over the flight, as higher request volume offsets the lower per-request cost. Multi-day engagement is flat across the flight ($N$-day retention for $N{\in}\{2,3,4,5\}$, all $|\Delta| \leq 0.10\%$, all $p > 0.45$), so the latency and efficiency gains do not come at the cost of engagement.
\fi

\paragraph{Flagged regressions.}
\ifanonymous
A small number of marginally significant scorecard deltas were investigated and traced to agent-layer failure modes and turn-1 model exploration rather than the routing decision; none was router-attributable.
\else
A small number of marginally significant scorecard deltas were investigated. Trajectory analysis attributed them to agent-layer failure modes (wrong edits, stalls, hallucinated paths) and to early model-exploration behavior rather than to model selection; none was router-attributable.
\fi

\section{Analysis}
\label{sec:analysis}

\ifanonymous
\paragraph{Dimension weights and frontier.}
Grid search yields non-uniform optimal weights---debugging and tool use outweigh reasoning and code generation, since cheap models handle routine code but struggle with subtle debugging and multi-step orchestration; the 50 largest quality-loss instances cluster into ambiguous intent (40\%), hidden complexity (35\%), and label noise (25\%). \textbf{HyDRA} traces a smooth, tunable Pareto curve on the 5-model SWE-Bench pool (Figure~\ref{fig:pareto}), from 87.0\% QR / 37.6\% CS at $\tau{=}0.10$ to 80.5\% QR / 78.5\% CS at $\tau{\geq}1.0$, while binary-v1 sits at a single dominated point (85.6\% QR / 9.1\% CS).
\else
\paragraph{Dimension weights and failure modes.}
Grid search reveals non-uniform optimal weights: debugging and tool use are weighted higher than reasoning and code generation---cheap models handle routine code well but struggle with subtle debugging and multi-step tool orchestration. The 50 largest quality-loss instances cluster into ambiguous intent (40\%), hidden complexity (35\%), and label noise (25\%).

\paragraph{Cost-quality frontier.}
\textbf{HyDRA} provides a smooth, continuously tunable Pareto curve on the 5-model SWE-Bench pool (Figure~\ref{fig:pareto}), ranging from 87.0\% QR / 37.6\% CS at $\tau{=}0.10$ to 80.5\% QR / 78.5\% CS at $\tau{\geq}1.0$. binary-v1 sits at a single dominated point (85.6\% QR / 9.1\% CS).
\fi

\newcommand{\figPareto}{%
\begin{figure}[t]
\centering
\begin{tikzpicture}
\begin{axis}[
    width=0.95\columnwidth,
    height=6cm,
    xlabel={Cost Savings vs.\ Sonnet (\%)},
    ylabel={Quality Retention vs.\ Oracle (\%)},
    xmin=0, xmax=85,
    ymin=78, ymax=92,
    grid=major,
    grid style={gray!30},
    legend pos=south west,
    legend style={font=\scriptsize},
    mark size=2.5pt,
]
\addplot[mark=*, blue, thick, mark options={fill=blue}] coordinates {
    (12.92, 87.47)
    (39.42, 87.24)
    (43.53, 86.54)
    (53.79, 86.08)
    (54.10, 85.85)
    (60.62, 84.92)
    (63.61, 84.22)
    (67.38, 83.99)
    (68.97, 83.06)
    (72.54, 82.37)
    (75.93, 81.67)
    (78.14, 80.97)
    (78.49, 80.51)
};
\addlegendentry{\textbf{HyDRA}}

\addplot[mark=square*, red, thick, mark options={fill=red}, only marks] coordinates {
    (9.07, 85.61)
};
\addlegendentry{binary-v1}

\addplot[mark=triangle*, gray, thick, mark options={fill=gray}, only marks] coordinates {
    (78.58, 80.51)
};
\addlegendentry{Always-Cheap}

\addplot[mark=diamond*, black, thick, mark options={fill=black}, only marks] coordinates {
    (0, 86.08)
};
\addlegendentry{Always-Strong}

\node[font=\tiny, anchor=south west] at (axis cs:12.92,87.47) {$\tau{=}0.01$};
\node[font=\tiny, anchor=south west] at (axis cs:54.10,85.85) {$\tau{=}0.24$};
\node[font=\tiny, anchor=south west] at (axis cs:72.54,82.37) {$\tau{=}0.64$};

\end{axis}
\end{tikzpicture}
\caption{Cost-quality Pareto frontier on SWE-Bench Verified (5-model pool). \textbf{HyDRA} offers a smooth, continuously tunable curve; binary-v1 provides only a single operating point that is dominated at every $\tau$.}
\label{fig:pareto}
\end{figure}
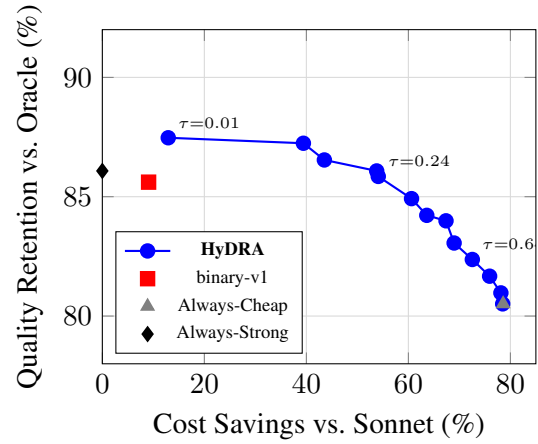%
}
\ifanonymous\else\figPareto\fi

\section{Production Deployment}
\label{sec:deployment}

\textbf{HyDRA} is deployed as a pre-routing layer on \gateway{}, serving all auto-mode traffic in \idechat{} for a large global developer population.

\ifanonymous
\paragraph{Serving and operations.}
\label{sec:stickiness}\label{sec:explainability}
\gatewayshort{} acts as a health \emph{veto} (fail-open), never a reranker. Adding, removing, or repricing a model is a YAML edit with zero retraining or downtime; at full rollout it meets availability targets at 55\,ms P50 overhead and steers most traffic to cheaper models. Serving (INT8 ONNX on CPU), image hard-gating, sticky routing, and quantization are in Appendices~\ref{app:deployment} and~\ref{app:explainability}.
\else
\paragraph{Integration and serving.} \gatewayshort{} operates as a health \emph{veto}, not a reranker: it removes unhealthy models from \textbf{HyDRA}'s ranked candidate list but never changes the ordering, with fail-open semantics guaranteeing availability. The deployed checkpoint is a dynamic INT8 ONNX model (attention nodes excluded from quantization) running on CPU. Image-bearing requests are routed only over vision-capable models: \textbf{HyDRA}'s candidate pool is pre-filtered to models with \texttt{vision:\ true} before shortfall matching, since the predictor is text-only by construction. A vision capability head is planned for a future model version (Appendix~\ref{app:image_hardgate}).

\paragraph{Prompt-cache-preserving sticky routing.}
\label{sec:stickiness}
Multi-turn agentic conversations span 20+ turns and accumulate tens of thousands of context tokens; provider prompt caches make this affordable at a 90\% discount, but switching models mid-conversation defeats the cache entirely. To protect the cache, the router is invoked in exactly three situations: (1)~turn~1 of a new conversation, (2)~after an explicit user-issued compaction, and (3)~after background summarization. Every other turn reuses the cached model, keyed on the conversation identifier so different conversations route independently. The service exposes three configurable stickiness modes---\textbf{per-request} (every turn re-routed), \textbf{per-content} (cache key includes a SHA-256 hash of the conversation prefix), and \textbf{per-session} (the production default).

\paragraph{Model lifecycle.}
\label{sec:lifecycle}
Adding, removing, or repricing a model requires only a YAML configuration edit (defining the capability profile via the benchmark-anchored computation of \S\ref{sec:profiles} and setting cost-per-token). Traffic redistributes automatically via shortfall matching---zero retraining, zero downtime.

\paragraph{Production deployment.}
\label{sec:server_metrics}
At full rollout the routing endpoint meets its availability targets, with routing overhead of 55\,ms P50 (under 1\% of end-to-end response time, \S\ref{sec:latency}). Shortfall matching steers the majority of traffic to the cheaper models in the pool. Full deployment architecture and quantization details are in Appendix~\ref{app:deployment}.

\paragraph{Routing explainability.}
\label{sec:explainability}
We have prototyped a developer-mode inspector in \idechat{} that surfaces the selected model and the four predicted capability scores per turn (Appendix~\ref{app:explainability}). This is an internal-only prototype at the time of writing.
\fi


\section{Conclusion}

\ifanonymous
\textbf{HyDRA} predicts \emph{what a query needs} and routes to \emph{the cheapest capable model}, decoupling the model catalog from the learned predictor so models are added, removed, or repriced with zero retraining---validated at scale by a near-million-user-per-arm production A/B flight.
\else
\textbf{HyDRA} reframes model routing as a two-part decision: predict \emph{what capabilities a query demands}, then select \emph{the cheapest model that meets them}. Only the first half is learned; the second is a configuration-defined capability profile, and this split fully decouples the router from the model catalog. New models are onboarded, and weaker ones retired, by editing a configuration file rather than retraining a classifier---over four months of deployment at \productname{} spanning six model additions and three removals, this required zero retraining. The decoupling does not come at the cost of quality: across SWE-Bench Verified, LiveCodeBench, and BigCodeBench, \textbf{HyDRA} holds quality within a fraction of a point of the strongest single model while cutting cost by more than half, and a controlled A/B flight with close to one million users per arm improved latency, time-to-first-token, reliability, and engagement with no statistically significant user-visible degradation in measured metrics, alongside an estimated 7--20\% reduction in serving cost-of-goods-sold for the routed segment.

The most immediate extensions are a vision capability dimension for multimodal routing, per-subagent capability prediction for agentic workflows, and adaptive multi-turn re-routing that balances prompt-cache preservation against mid-conversation complexity drift. More broadly, we see capability-based, catalog-decoupled routing as a practical foundation for operating the heterogeneous, fast-churning model pools that production LLM systems increasingly depend on.
\fi

\section*{Limitations}
\label{sec:limitations}

\textbf{Domain and deployment scope.} The four dimensions target coding tasks; other domains require redefining dimensions and relabeling (the architecture generalizes). Our production measurements come from a single surface, \idechat{}; because routing is fully model- and surface-decoupled, rollout to \siblingsurfaces{} is underway and requires no retraining. Our cross-benchmark results (\S\ref{sec:exp_crossbench}) show the capability dimensions and shortfall policy transfer across off-distribution coding \emph{tasks}, but generalization across \emph{surfaces} (which differ in traffic mix and interaction modality) and to non-coding domains has not yet been measured at production scale.

\textbf{Judge bias and repo-dependent queries.} Position swapping mitigates positional bias; other biases (verbosity, sycophancy) remain. Telemetry queries are stripped of repository state before labeling, so the LLM judge cannot fully resolve repository-dependent queries (``why does this test fail''), biasing scores toward high-requirement defaults and limiting cost savings on this slice.

\textbf{Reliance on a moderate-agreement judge.} Both our training labels and the QR metric derive from an LLM judge whose \emph{absolute} agreement with adjudicated human labels is modest (pooled Krippendorff's $\alpha{=}0.24$). Three factors limit the impact on routing: (i)~the router acts on \emph{relative} per-dimension requirement bands calibrated to \textbf{HyDRA}'s own score distribution rather than on absolute judge scores, so a uniform judge bias is largely absorbed by profile calibration; (ii)~against the human audit set \textbf{HyDRA} tracks adjudicated labels better than the judge it distills from (Krippendorff's $\alpha{=}0.40$ vs.\ $0.24$, MAE $0.15$ vs.\ $0.23$; Appendix~\ref{app:human_eval}); and (iii)~end-to-end quality is corroborated by two judge-independent signals---SWE-Bench Verified resolution and the production A/B flight. Conclusions that hinge on absolute judge scores should nonetheless be read with this agreement level in mind.

\textbf{Benchmark-derived profiles.} Public benchmarks may not reflect real query performance; profile errors cause systematic misrouting. SWE-Bench is reasoning-heavy---real traffic has more trivial queries, likely increasing savings.

\textbf{Multilingual coverage imbalance.} CJK and ``other'' language groups have fewer training samples (250 and ${\sim}$300 conversations) than European (42K), so per-language evaluation may be underpowered for low-resource languages.

\textbf{Tool-use calibration.} Tool use is highly weighted in routing (\S\ref{sec:analysis}) yet shows the lowest agreement with human labels (Krippendorff's $\alpha{=}{-}0.04$, essentially chance-level): a near-constant under-prediction offset on non-English queries, not a rank inversion. Because profiles are calibrated on \textbf{HyDRA}'s own score band this offset is largely absorbed---end-to-end quality holds on SWE-Bench and the production A/B flight---and is correctable by affine recalibration (future work for non-English traffic); see Appendix~\ref{app:human_eval}.

\textbf{Adversarial robustness.} Recent work~\citep{tang2026r2a} shows adversarial suffix optimization can manipulate LLM routers; Appendix~\ref{app:adversarial_probe} reports a diagnostic probe for \textbf{HyDRA}. Defenses (input perturbation detection, suffix filtering, score anomaly filtering, calibrated score caps) remain future work.

\textbf{Compaction-bounded re-routing and context cap.} The router is invoked on the first turn and after each compaction event; between compactions the selected model is reused via prompt-cache--preserving sticky routing (\S\ref{sec:stickiness}), so mid-conversation difficulty drift cannot trigger re-routing. The deployed checkpoint also truncates inputs at 512 tokens, which can underestimate complexity on long contexts.

\textbf{Labeling cost.} Dual-model generation + LLM judging dominates labeling expense; we do not report a precise dollar figure because runs combine on-demand and \ifanonymous batch-API\else Azure Batch\fi{} pricing across multiple regions.

\section*{Ethics Statement}
\label{sec:ethics}

\textbf{Data sourcing and consent.} The data used for training and evaluation (\S\ref{sec:labeling}) consists of de-identified, user-provided prompts and interaction content drawn from \datasrc{} for users who have opted in to product-improvement data sharing under \productname{}'s privacy and data-handling agreement. No third-party or publicly scraped datasets are used. The dataset does not include private repository contents beyond what users explicitly provide in their prompts, and is subject to \productname{}'s standard content-filtering and de-identification controls: personally identifiable information is removed prior to labeling and storage, and raw user prompts are not redistributed. All experiments, including the production A/B flight (\S\ref{sec:exp_abtest}), were conducted in accordance with \productname{}'s applicable terms of service and experimentation policies, and users retain the ability to select models manually or opt out of auto-routing.

\textbf{Annotator compensation.} The native-speaker annotators contracted for the human-labeled multilingual eval set (Appendix~\ref{app:human_eval}) were compensated at the contracting vendor's standard rate for technical-content annotation in their respective markets. Annotators were briefed on the labeling rubric and the downstream routing use case prior to consent.

\textbf{LLM judge bias.} Quality labels are produced by an LLM-as-judge pipeline (\S\ref{sec:labeling}). We mitigate positional bias via order swapping but cannot fully eliminate verbosity, style, or self-preference biases. The held-out human-labeled eval set (Appendix~\ref{app:human_eval}) provides an independent human-consensus audit of the deployed requirement predictor.

\textbf{Cost--quality trade-off and end-user impact.} \textbf{HyDRA} routes auto-mode requests to cheaper models when the predicted capability requirement allows. This trades a small expected-quality reduction for substantial cost savings on a population that did not explicitly opt in to a specific model. We mitigate this by reporting per-language quality-retention numbers and per-SKU cost outcomes, surfacing the routing decision and per-dimension scores in a developer-mode inspector (\S\ref{sec:explainability}, currently internal-only), and maintaining a user-facing model-picker override path so any user can opt out of auto-routing entirely.

\textbf{Security and adversarial use.} The adversarial suffix probe in Appendix~\ref{app:adversarial_probe} documents a known cost-inflation attack surface. We disclose this openly so deployers can plan defenses; we are not aware of in-the-wild exploitation against the deployed router.

\ifanonymous\else
\section*{Acknowledgments}

We thank the \productname{}, \ideapp{}, and \gatewayshort{} infrastructure teams for the production integration, telemetry, and rollout support that made this work possible, and our annotation partners for the multilingual labeling effort.
\fi

\raggedbottom
\bibliography{hydra_refs}

\appendix
\flushbottom

\ifanonymous
\tabCompMatched
\figParetoCompare
\figPareto
\fi

\section{Competitive Design Comparison}
\label{app:competitive_design}

Table~\ref{tab:competitive} situates \textbf{HyDRA} against prior LLM routing systems along the design axes that matter for production deployment.

\begin{table*}[!htbp]
\centering
\resizebox{\textwidth}{!}{%
\begin{tabular}{lcccccc}
\toprule
\textbf{System} & \textbf{Routing} & \textbf{Model-} & \textbf{\# Dims} & \textbf{Latency} & \textbf{Retraining on} & \textbf{Multi-} \\
 & \textbf{Type} & \textbf{Decoupled?} & & \textbf{Overhead} & \textbf{Catalog Change?} & \textbf{Lingual?} \\
\midrule
RouteLLM~\citep{ong2024routellm} & Pre-route & No & 1 (binary) & $\sim$50ms & Yes & No \\
Hybrid LLM~\citep{ding2024hybrid} & Pre-route & No & 1 (scalar) & $\sim$40ms & Yes & No \\
FrugalGPT~\citep{chen2023frugalgpt} & Cascade & No & 1 & $+$latency/step & Yes & No \\
AutoMix~\citep{madaan2024automix} & Cascade+verify & No & 1 & $+$verify cost & Yes & No \\
ZOOTER~\citep{lu2024routing} & Pre-route & Partial & 1 & $N\times$reward & No & No \\
EcoAssistant~\citep{zhang2024ecoassistant} & Cascade+cache & No & 1 & $+$cache lookup & Yes & No \\
MTRouter~\citep{zhang2026mtrouter} & Pre-route & No & 1 & $\sim$50ms & Yes & No \\
TRouter~\citep{liu2026trouter} & Pre-route & No & 1 & $\sim$40ms & Yes & No \\
DialRouter~\citep{zhang2026dialrouter} & Pre-route & No & 1 & $+$MCTS & Yes & No \\
LLM Router~\citep{varshney2026llmrouter} & Pre-route & No & 1 & $+$prefill & Yes & No \\
Avengers Pro~\citep{zhang2025avengerspro} & Pre-route & No & 1 (scalar) & $\sim$40ms & Yes & No \\
\midrule
\textbf{HyDRA} (ours) & Pre-route & \textbf{Yes} & \textbf{4} & \textbf{55ms P50} & \textbf{No} & \textbf{Yes} \\
\bottomrule
\end{tabular}%
}
\caption{Competitive analysis of LLM routing systems. \textbf{HyDRA} is the only system that is fully model-decoupled (no retraining on catalog change), uses multi-dimensional capability prediction, and provides multilingual routing across 16 languages.}
\label{tab:competitive}
\end{table*}

\begin{table}[!htbp]
\centering
\small
\setlength{\tabcolsep}{4pt}
\begin{tabularx}{\linewidth}{@{}>{\raggedright\arraybackslash}X r r r r@{}}
\toprule
\textbf{Router (threshold)} & \textbf{Res\%} & \textbf{QR} & \textbf{CS} & \textbf{Mis.} \\
\midrule
RouteLLM (BERT, $t{=}0$)   & 73.40 & 100.00 & +0.00  & 69.40 \\
RouteLLM (BERT, $t{=}25$)  & 73.40 & 100.00 & +0.44  & 69.20 \\
RouteLLM (BERT, $t{=}50$)  & 72.00 & 98.09  & +18.57 & 45.20 \\
RouteLLM (BERT, $t{=}75$)  & 69.40 & 94.55  & +49.69 & \phantom{0}0.40 \\
RouteLLM (BERT, $t{=}100$) & 69.40 & 94.55  & +49.86 & \phantom{0}0.00 \\
\midrule
\textbf{HyDRA} (cons., $\tau{=}1.15$) & 72.80 & 99.18 & +20.20 & 40.60 \\
\textbf{HyDRA} (agg., $\tau{=}1.42$)  & 69.80 & 95.10 & +44.50 & \phantom{0}7.40 \\
\bottomrule
\end{tabularx}
\caption{RouteLLM (BERT) threshold sweep on SWE-Bench Verified, evaluated on a 2-model pool (strong = GPT-5.3 Codex, weak = GPT-5.4 mini); higher threshold $\to$ easier to choose weak. RouteLLM only supports binary (2-model) routing, which is why this comparison is restricted to a strong/weak pair rather than the 5-model pool used elsewhere in the paper.}
\label{tab:routellm_comparison}
\end{table}

\begin{table*}[!htbp]
\centering
\small
\begin{tabularx}{\textwidth}{@{}>{\raggedright\arraybackslash}p{3.4cm} c >{\raggedright\arraybackslash}X c >{\raggedright\arraybackslash}X@{}}
\toprule
 & \multicolumn{2}{c}{\textbf{HyDRA}} & \multicolumn{2}{c}{\textbf{binary-v1}} \\
\cmidrule(lr){2-3}\cmidrule(lr){4-5}
\textbf{Catalog event} & \textbf{QR} & \textbf{Adapts?} & \textbf{QR} & \textbf{Adapts?} \\
\midrule
Original 3-model pool                  & 98.1\%             & ---                                & baseline   & --- \\
Remove Claude Haiku 4.5                & 95.7\% ($-$2.4)    & Yes (YAML, no retrain)             & unchanged  & No (model never used) \\
Add new mid-tier model                 & 98.5\% ($+$0.4)    & Yes (routes to new model day 1)    & unchanged  & No (cannot use new model) \\
\bottomrule
\end{tabularx}
\caption{\textbf{Catalog portability.} \textbf{HyDRA} reroutes via a YAML edit: a small QR drop on removal, a small QR gain on addition. binary-v1 cannot perceive catalog changes without retraining.}
\label{tab:portability}
\end{table*}

\section{Algorithm Pseudocode}
\label{app:algorithms}

\begin{algorithm}[t]
\caption{Capability Profile Computation}
\label{alg:capability}
\begin{algorithmic}[1]
\REQUIRE Benchmark scores $\{s_{m,b}\}$ for models $m \in \mathcal{M}$, benchmarks/subgroups $b$
\REQUIRE Benchmark/subgroup importance weights $\{\alpha_b\}$ and judge dimension weights $\{\omega_{b,k}\}$ ($\omega_{b,k}{=}0$ when benchmark $b$ does not exercise dimension $k$, i.e.\ $b \notin \mathcal{B}(k)$; Table~\ref{tab:capability_weights})
\REQUIRE Operator-stored routing weights $\{w_k\}$ (per dimension)
\ENSURE Capability profiles $\{c_{m,k}\}$ and band-compensated routing weights $\{\tilde{w}_k\}$
\FOR{each dimension $k$ and each model $m \in \mathcal{M}$}
  \STATE $\text{raw}_{m,k} \leftarrow \dfrac{\sum_{b} \alpha_b\, \omega_{b,k}\, s_{m,b}}{\sum_{b} \alpha_b}$ \hfill \COMMENT{Step 1: benchmark anchoring}
\ENDFOR
\STATE $[\beta^{\text{lo}}_k, \beta^{\text{hi}}_k] \leftarrow$ low/high percentiles of the requirement predictor's score distribution on a held-out set, per dimension $k$
\FOR{each dimension $k$ and each model $m \in \mathcal{M}$}
  \STATE $c_{m,k} \leftarrow
          \beta^{\text{lo}}_k +
          \dfrac{\text{raw}_{m,k} - \min_j \text{raw}_{j,k}}
                {\max_j \text{raw}_{j,k} - \min_j \text{raw}_{j,k}}
          \cdot (\beta^{\text{hi}}_k - \beta^{\text{lo}}_k)$ \hfill \COMMENT{Step 2: pool-relative normalization}
\ENDFOR
\STATE $\Delta_k \leftarrow \beta^{\text{hi}}_k - \beta^{\text{lo}}_k$ \hfill \COMMENT{Step 3: band-width dim-weight compensation (Eq.~\ref{eq:dimcomp})}
\STATE $\tilde{w}_k \leftarrow \dfrac{w_k / \Delta_k}{\sum_{k'} w_{k'} / \Delta_{k'}} \cdot \sum_{k'} w_{k'}$
\end{algorithmic}
\end{algorithm}

\begin{algorithm}[t]
\caption{Shortfall-Based Routing Algorithm}
\label{alg:routing}
\begin{algorithmic}[1]
\REQUIRE Requirements $\hat{\mathbf{r}} = (\hat{r}_1, \ldots, \hat{r}_K)$, threshold $\tau$
\REQUIRE Model pool $\{(m, \mathbf{c}_m, \text{cost}_m)\}$, band-compensated weights $\tilde{\mathbf{w}} = (\tilde{w}_1, \ldots, \tilde{w}_K)$ from Eq.~\ref{eq:dimcomp}
\REQUIRE Available models $\mathcal{A}$ (from infrastructure health filter)
\REQUIRE Routing confidence $\gamma = \max_k \hat{r}_k$, sticky threshold $\gamma_{\text{sticky}}$
\ENSURE Selected model $m^*$
\IF{$\gamma < \gamma_{\text{sticky}}$}
  \RETURN first model in $\mathcal{A}$ \hfill \COMMENT{sticky: keep current model}
\ENDIF
\STATE $\mathcal{F} \leftarrow \text{PreFilter}(\mathcal{A})$ \hfill \COMMENT{hard gates (e.g., vision)}
\FOR{each $m \in \mathcal{F}$}
  \STATE $s_m \leftarrow \sum_{k=1}^{K} \tilde{w}_k \cdot \max(0, \hat{r}_k - c_{m,k})$
\ENDFOR
\STATE $\mathcal{E} \leftarrow \{m \in \mathcal{F} : s_m \leq \tau\}$ \hfill \COMMENT{eligible set}
\IF{$\mathcal{E} \neq \emptyset$}
  \STATE $m^* \leftarrow \arg\min_{m \in \mathcal{E}} \text{cost}_m$ \hfill \COMMENT{cheapest eligible}
\ELSE
  \STATE $m^* \leftarrow \arg\min_{m \in \mathcal{F}} s_m$ \hfill \COMMENT{fail-open: least shortfall}
\ENDIF
\RETURN $m^*$
\end{algorithmic}
\end{algorithm}

\begin{algorithm}[t]
\caption{Router Evaluation Metrics Computation}
\label{alg:metrics}
\begin{algorithmic}[1]
\REQUIRE Queries $\mathcal{Q}$, per-model outcomes $\{y_{q,m}\}$, costs $\{\text{cost}_{q,m}\}$
\REQUIRE Router assignments $\{m_q\}$, baseline model $m_{\text{base}}$
\ENSURE $\text{QR}, \text{CS}, \text{Mis}$
\STATE $\text{res}_r \leftarrow \sum_{q} y_{q, m_q}$; \enspace $\text{cost}_r \leftarrow \sum_{q} \text{cost}_{q, m_q}$
\STATE $\text{cost}_b \leftarrow \sum_{q} \text{cost}_{q, m_{\text{base}}}$
\STATE order models $m_{(1)} \prec \cdots \prec m_{(|M|)}$ by input/output unit prices \hfill \COMMENT{static price-card order}
\STATE $\text{rank}(m) \leftarrow$ position of $m$ in this ordering
\STATE $n_{\text{oracle}} \leftarrow 0$;\enspace $n_{\text{mis}} \leftarrow 0$
\FOR{each $q \in \mathcal{Q}$}
  \STATE $\mathcal{R}_q \leftarrow \{\, m : y_{q,m}=1 \,\}$
  \IF{$\mathcal{R}_q \neq \emptyset$}
    \STATE $n_{\text{oracle}} \leftarrow n_{\text{oracle}} + 1$
    \IF{$\exists\, m \in \mathcal{R}_q : \text{rank}(m) < \text{rank}(m_q)$}
      \STATE $n_{\text{mis}} \leftarrow n_{\text{mis}} + 1$ \hfill \COMMENT{a globally cheaper model resolved $q$}
    \ENDIF
  \ENDIF
\ENDFOR
\STATE $\text{QR} \leftarrow \text{res}_r / n_{\text{oracle}}$
\STATE $\text{CS} \leftarrow 1 - \text{cost}_r / \text{cost}_b$
\STATE $\text{Mis} \leftarrow n_{\text{mis}} / |\mathcal{Q}|$
\end{algorithmic}
\end{algorithm}

\section{Deployment Details}
\label{app:deployment}

This appendix provides the full deployment architecture, quantization details, and model lifecycle operations summarized in \S\ref{sec:deployment}. The sticky routing policy is described in \S\ref{sec:stickiness}.

\subsection{\gatewayshort{} Integration Architecture}
\label{app:capi}

The existing \gatewayshort{} routing infrastructure makes model selection decisions based on infrastructure health metrics---throughput utilization, error rates, and latency---using weight multipliers per model endpoint. This is entirely \emph{content-blind}. \textbf{HyDRA} integrates via a \textbf{rank-then-filter} protocol (Algorithm~\ref{alg:capi}):

\begin{algorithm}[!ht]
\caption{\gatewayshort{} Integration: Rank-then-Filter}
\label{alg:capi}
\begin{algorithmic}[1]
\REQUIRE Query $q$, available models $\mathcal{A}$ (from \gatewayshort{})
\REQUIRE Health scores $\{h_m\}$, health floor $h_f = 0.10$
\ENSURE Selected model $m^*$
\STATE $\hat{\mathbf{r}}, \gamma \leftarrow \text{HyDRA}.\text{predict}(q)$
\STATE $\mathcal{E} \leftarrow \text{ShortfallMatch}(\hat{\mathbf{r}}, \mathcal{A})$ \hfill \COMMENT{Alg.~\ref{alg:routing}}
\STATE $\mathcal{H} \leftarrow \{m \in \mathcal{E} : h_m \geq h_f\}$ \hfill \COMMENT{health veto}
\IF{$\mathcal{H} = \emptyset$}
  \RETURN $\text{\gatewayshort{}\_Fallback}(\mathcal{A}, \{h_m\})$ \hfill \COMMENT{fail-open}
\ENDIF
\RETURN $\text{first}(\mathcal{H})$ \hfill \COMMENT{cheapest capable \& healthy}
\end{algorithmic}
\end{algorithm}

Three design properties govern this integration: (1)~\gatewayshort{} is a veto, not a reranker---it removes unhealthy models but never changes the ordering; (2)~fail-open semantics guarantee availability when all eligible models are unhealthy; (3)~\gatewayshort{}'s stochastic weighted sampling is replaced by deterministic cheapest-eligible selection, making routing reproducible.

\subsection{Image Hardgating}
\label{app:image_hardgate}

When the incoming request contains image attachments, the available-model list is pre-filtered to vision-capable models (those with \texttt{vision:\ true} in the capability profile) before \textbf{HyDRA} performs shortfall matching; if no vision-capable model is available, the router returns 400 so the caller can fall back. \textbf{HyDRA} still produces capability requirement scores from the text portion of the request---only the candidate pool is restricted. This guard is in place because (i)~the training distribution contains no image content and (ii)~the predictor lacks a calibrated vision dimension. Adding a vision capability head---and the OCR / visual-complexity signals needed to train it---is on the roadmap for a future model version.

\subsection{Quantization and Inference}
\label{app:quantization}

The deployed checkpoint is exported from PyTorch to ONNX FP32, then quantized in two variants:

\textbf{Dynamic INT8 with attention nodes excluded} (production). ONNX Runtime's dynamic INT8 weight quantization is applied while excluding all attention/QKV nodes. Full INT8 quantization of ModernBERT collapses predictions because the attention layer's masked-fill operations interact pathologically with INT8 calibration; excluding attention nodes preserves accuracy while cutting model size by ${\sim}$40\% and improving CPU latency by ${\sim}$8\%.

\textbf{FP16 ONNX} via float-to-float16 conversion. Used on hardware where FP16 kernels are faster than dynamic INT8.

We also implement \emph{quantization-aware training} (QAT) as a fallback for future encoders or platforms where dynamic INT8 fails; it was not required for the deployed checkpoint.

\section{In-Product Routing Explainability Prototype}
\label{app:explainability}

We have prototyped a developer-mode routing inspector for \idechat{}'s auto-routing surface (an \emph{internal mock; not yet shipped to end users}). When enabled, the inspector surfaces the model \textbf{HyDRA} selected alongside the four sigmoid-head requirement scores (reasoning, code generation, debugging, tool use) that drove the shortfall-matching decision, making an otherwise opaque routing choice legible to the developer. Because the capability scores are produced as a byproduct of every routing decision, exposing them carries no additional inference cost.

\ifanonymous\else
\begin{figure}[t]
  \centering
  \includegraphics[height=0.36\textheight,keepaspectratio]{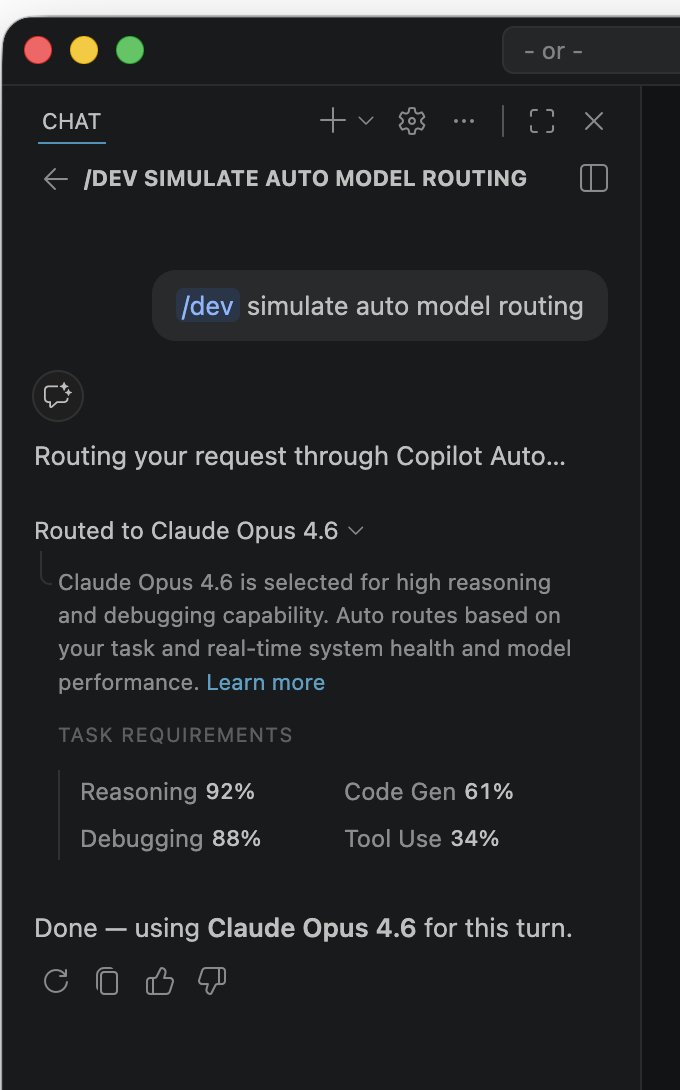}
  \caption{Prototype routing explainability inspector for \idechat{}'s developer-mode auto-routing surface (\emph{internal mock; not yet shipped to end users}). The chosen model is displayed alongside the four sigmoid-head outputs that drove the shortfall-matching decision (here: Reasoning 92\%, Code Generation 61\%, Debugging 88\%, Tool Use 34\%).}
  \label{fig:vscode_explainability}
\end{figure}
\fi

\section{Human-Labeled Multilingual Eval Set}
\label{app:human_eval}

In addition to the LLM-judge--labeled training and eval sets (\S\ref{sec:labeling}, \S\ref{sec:exp_telemetry}), we curate a separate \textbf{human-labeled multilingual eval set} drawn from \datasrc{} telemetry. This set provides an independent reference for auditing the deployed requirement predictor against human judgments rather than only against LLM-generated supervision.

\paragraph{Composition.}
The audit set comprises 3{,}819 tasks spanning 16 languages (English plus 15 non-English buckets in the CJK / European / Other groups, following the production-eval language distribution). Every task carries three independent per-dimension rater scores; metrics are computed on the first user turn of the non-compacted subset. All tasks are drawn from the same \datasrc{} telemetry distribution used in the multilingual production eval.

\paragraph{Annotation protocol.}
Each query is independently labeled by \textbf{three annotators} drawn from vendor-supplied annotation teams with native-speaker coverage of the language groups; all annotators completed a 200-example calibration pass against an internal gold set before live labeling. Annotators score the query along the same four capability dimensions (reasoning, code generation, debugging, tool use) on the $1$--$5$ rubric used by the LLM judge, plus a free-text rationale. A senior reviewer then adjudicates the three independent rater scores into a single per-dimension human reference (the adjudicated reference), resolving disagreements case-by-case under the same rubric rather than mechanically averaging the raters or relying on any vendor-supplied ``final'' label. The adjudicator is a domain expert who reconciles---rather than overrides---the raters, using their scores and rationales as evidence; a single adjudicated reference trades the variance-reduction of rater averaging for consistency, at the cost of a single point of judgment, which we treat as acceptable for a held-out audit that informs no model-selection decision.

\paragraph{Use in evaluation.}
We use this set strictly as a held-out audit: the deployed checkpoint was selected before these labels were available, and no model-selection or threshold-tuning decision uses them. We take the senior-adjudicated score as the primary human reference and compare three quantities: agreement among the human raters, agreement between the human reference and \textbf{HyDRA}'s continuous requirement scores, and agreement between the human reference and a gpt-5.2 single-response LLM judge (Table~\ref{tab:human_alpha}). Table~\ref{tab:mae_human} reports the corresponding absolute error after mapping all scores to the shared $[0,1]$ scale.

\paragraph{Agreement and MAE.}
Human annotators show moderate-to-high interval agreement ($\alpha=0.477$--$0.627$ per dimension; pooled $\alpha=0.635$), indicating that the rubric is usable but still noisy at the per-dimension level. Against the human reference, \textbf{HyDRA} has higher pooled agreement than the LLM judge ($\alpha=0.402$ vs.\ $0.235$) and lower pooled MAE ($0.154$ vs.\ $0.228$): the deployed predictor tracks the adjudicated human labels at least as closely as a strong single-response judge on this slice. The one clear exception is \textbf{tool use}: humans agree strongly with one another ($\alpha=0.627$), but \textbf{HyDRA}'s interval agreement with the human reference is essentially zero ($\alpha=-0.039$). This is a calibration gap rather than an inversion---\textbf{HyDRA}'s tool-use scores are weakly but positively rank-correlated with the human reference (Spearman $\rho=+0.22$) but sit on average ${\approx}0.15$ below it on the $[0,1]$ scale (human mean $0.28$ vs.\ predictor mean $0.13$), so the negative $\alpha$ reflects a roughly constant under-prediction offset rather than a reversal of which tasks require tools. The gap is concentrated in non-English queries: partitioning the audit set by language, tool-use agreement is positive on the English subset ($\alpha=+0.081$, $n=2{,}474$) and negative on the non-English subset ($\alpha=-0.205$, $n=1{,}345$), localizing the residual to absolute calibration on non-English tool-use queries; we discuss the routing implications below. The LLM judge is evaluated with a compact single-response rendering that includes the user prompt, assistant text, thinking text, and tool-call names, but excludes tool inputs and tool results.

\begin{table}[!htbp]
\centering
\footnotesize
\setlength{\tabcolsep}{3pt}
\resizebox{\columnwidth}{!}{%
\begin{tabular}{lccc}
\toprule
\textbf{Dim.} & \textbf{Inter-ann.} & \textbf{Human vs HyDRA} & \textbf{Human vs judge} \\
\midrule
Reasoning & 0.477 & 0.460 & $-$0.068 \\
Code Gen  & 0.524 & 0.308 & $-$0.023 \\
Debugging & 0.601 & 0.525 & 0.256 \\
Tool Use  & 0.627 & $-$0.039 & 0.245 \\
\midrule
\textit{Pooled} & 0.635 & 0.402 & 0.235 \\
\bottomrule
\end{tabular}
}
\caption{Interval Krippendorff's $\alpha$ on the human-labeled audit set (3{,}819 tasks across 16 languages; turn 0, non-compacted). The first column is inter-annotator agreement among three human raters on the raw 1--5 scale; the remaining columns compare the senior-adjudicated human reference (rescaled to $[0,1]$) with \textbf{HyDRA}'s continuous scores and the gpt-5.2 judge scores. The pooled row is one $\alpha$ over stacked $(task, dim)$ cells, not an average of the four per-dim values.}
\label{tab:human_alpha}
\end{table}

\begin{table}[!htbp]
\centering
\footnotesize
\begin{tabular}{lcc}
\toprule
\textbf{Dim.} & \textbf{HyDRA MAE} & \textbf{Judge MAE} \\
\midrule
Reasoning & 0.159 & 0.296 \\
Code Gen  & 0.149 & 0.195 \\
Debugging & 0.089 & 0.216 \\
Tool Use  & 0.217 & 0.204 \\
\midrule
\textit{Pooled} & 0.154 & 0.228 \\
\bottomrule
\end{tabular}
\caption{Per-dimension MAE on the shared $[0,1]$ scale against the senior-adjudicated human reference on the same audit set as Table~\ref{tab:human_alpha}.}
\label{tab:mae_human}
\end{table}

\paragraph{Routing implications.}
Two properties bound the effect of the tool-use offset on routing. First, capability profiles and dimension weights are calibrated onto \textbf{HyDRA}'s own empirical score band (\S\ref{sec:profiles}), so a roughly constant per-dimension offset shifts all candidates together and is largely absorbed by the operating point rather than changing which model is selected; accordingly, routing quality is established directly by end-to-end metrics---within $0.3$ pp at $54.1\%$ savings on SWE-Bench (\S\ref{sec:exp_swebench}) and no statistically significant user-visible degradation in measured metrics in the production A/B flight (\S\ref{sec:exp_abtest})---not inferred from this audit $\alpha$. Second, because the residual is a level bias rather than a rank inversion, it is correctable by per-dimension affine recalibration; restoring absolute tool-use calibration on non-English traffic, where the offset concentrates, and strengthening the still-weak positive rank correlation are future work. We note a structural tension worth flagging: tool use simultaneously carries the \emph{largest} band-compensated routing weight ($\tilde{w}\,{\approx}\,1.29$; Table~\ref{tab:profiles_summary}) and the \emph{weakest} human-agreement calibration. These are not in conflict by accident---the weight is set by inverse band width (Eq.~\ref{eq:dimcomp}), so the same narrow tool-use score band that drives the constant offset also inflates the dimension's weight, independently of its reliability. The band-calibration argument above is therefore load-bearing for tool use specifically, and revisiting the tool-use weight (e.g.\ down-weighting until non-English calibration is restored) is a concrete future-work lever.

\section{Extended Evaluation}
\label{app:extended_eval}

Additional evaluation results supporting the main-paper claims.

\begin{table}[H]
\centering
\small
\begin{tabular}{lccc}
\toprule
\textbf{Model} & \textbf{Res.\ \%} & \textbf{Input \$/M} & \textbf{Output \$/M} \\
\midrule
GPT-5             & 67.00 & \$0.84  & \$10.00 \\
GPT-5-mini        & 57.40 & \$0.20  & \$2.00  \\
GPT-5.2           & 73.20 & \$0.75  & \$14.00 \\
GPT-5.3-Codex     & 73.40 & \$0.58  & \$14.01 \\
GPT-5.4           & 73.40 & \$1.35  & \$15.22 \\
GPT-5.4-mini      & 69.40 & \$0.31  & \$4.50  \\
Claude-Sonnet-4.6 & 74.20 & \$1.19  & \$15.00 \\
Claude-Haiku-4.5  & 69.40 & \$0.52  & \$5.00  \\
\bottomrule
\end{tabular}
\caption{8-model offline evaluation pool used across all tracks in \S\ref{sec:experiments}. Res.\ \% is SWE-Bench Verified resolution rate; prices are per million input/output tokens.}
\label{tab:model_pool}
\end{table}

\subsection{Statistical Significance on SWE-Bench Verified}
\label{app:significance}

SWE-Bench Verified scores every router on the same 500 instances, so resolution-rate differences admit paired tests. For each router we report (i)~the 95\% Wilson interval on its resolution rate, (ii)~a 95\% paired-bootstrap interval on cost savings ($10{,}000$ resamples), and (iii)~a McNemar paired test of its per-instance resolve/no-resolve pattern against a reference router. All three are computed from the same per-instance resolution outcomes underlying Tables~\ref{tab:main} and~\ref{tab:competitive_matched}, so the point estimates here match those tables exactly. The headline reading: \textbf{HyDRA}'s quality is statistically indistinguishable from always-strong while its cost savings are large and exclude zero, and on the matched pool \textbf{HyDRA} is statistically tied with the strongest competitor at every operating point.

\begin{table*}[t]
\centering
\small
\textbf{(a) 5-model SWE-Bench pool} (companion to Table~\ref{tab:main}; $n{=}500$, Oracle resolves 431)\\[2pt]
\begin{tabular}{lcccccc}
\toprule
\textbf{Router} & \textbf{Res.\ \%} & \textbf{95\% CI} & \textbf{QR} & \textbf{CS \%} & \textbf{95\% CI} & \textbf{$p^{\dagger}$} \\
\midrule
\rowcolor{blue!10}
\textit{Oracle Routing} & \textit{86.2} & \textit{$[82.9, 88.9]$} & \textit{100.0} & \textit{---} & \textit{---} & \textit{---} \\
\rowcolor{green!12}
\textit{GPT-5.4-mini (cheap)} & \textit{69.4} & \textit{$[65.2, 73.3]$} & \textit{80.5} & \textit{78.6} & \textit{$[74.3, 82.1]$} & \textit{0.011} \\
Claude Haiku 4.5 & 69.4 & $[65.2, 73.3]$ & 80.5 & 11.7 & $[-6.3, 26.5]$ & 0.008 \\
GPT-5.3 Codex & 73.4 & $[69.4, 77.1]$ & 85.2 & 57.3 & $[48.9, 64.1]$ & 0.724 \\
GPT-5.4 & 73.4 & $[69.4, 77.1]$ & 85.2 & 25.7 & $[11.3, 37.4]$ & 0.720 \\
\rowcolor{gray!15}
\textit{Claude Sonnet 4.6 (strong)} & \textit{74.2} & \textit{$[70.2, 77.8]$} & \textit{86.1} & \textit{0.0} & \textit{---} & \textit{ref} \\
\midrule
\textbf{HyDRA (peak, $\tau{=}0.01$)} & \textbf{75.4} & $[71.4, 79.0]$ & \textbf{87.5} & \textbf{12.9} & $[5.3, 22.3]$ & 0.238 \\
\textbf{HyDRA (cons., $\tau{=}0.24$)} & \textbf{74.0} & $[70.0, 77.7]$ & \textbf{85.8} & \textbf{54.1} & $[44.8, 61.7]$ & 1.000 \\
\textbf{HyDRA (agg., $\tau{=}0.64$)} & \textbf{71.0} & $[66.9, 74.8]$ & \textbf{82.4} & \textbf{72.5} & $[67.0, 77.0]$ & 0.085 \\
\bottomrule
\end{tabular}\\[7pt]
\textbf{(b) Matched 3-model pool} (companion to Table~\ref{tab:competitive_matched}; $n{=}500$)\\[2pt]
\begin{tabular}{lcccccc}
\toprule
\textbf{System} & \textbf{Res.\ \%} & \textbf{95\% CI} & \textbf{QR} & \textbf{CS \%} & \textbf{95\% CI} & \textbf{$p^{\dagger}$} \\
\midrule
\rowcolor{gray!15}
\textit{GPT-5.2 (always-strong)} & \textit{73.2} & \textit{$[69.2, 76.9]$} & \textit{93.4} & \textit{0.0} & \textit{---} & \textit{ref} \\
Avengers Pro (Cons.) & 72.0 & $[67.9, 75.8]$ & 91.8 & 15.6 & $[12.0, 19.5]$ & 0.238 \\
OpenRouter Auto (c/q${=}0$) & 70.8 & $[66.7, 74.6]$ & 90.3 & \phantom{0}4.9 & $[1.5, 8.2]$ & 0.012 \\
Azure Foundry (Quality) & 65.0 & $[60.7, 69.1]$ & 82.9 & 36.3 & $[30.7, 41.8]$ & $<$0.001 \\
OpenRouter Auto (c/q${=}1$) & 64.4 & $[60.1, 68.5]$ & 82.1 & 44.2 & $[38.8, 49.4]$ & $<$0.001 \\
Avengers Pro (Aggr.) & 64.0 & $[59.7, 68.1]$ & 81.6 & 46.7 & $[41.9, 51.4]$ & $<$0.001 \\
Azure Foundry (Balanced) & 59.6 & $[55.2, 63.8]$ & 76.0 & 66.2 & $[61.8, 70.5]$ & $<$0.001 \\
\midrule
\textbf{HyDRA (cons., $\tau{=}0.703$)} & \textbf{70.8} & $[66.7, 74.6]$ & \textbf{90.3} & \textbf{16.2} & $[12.7, 20.0]$ & 0.004 \\
\textbf{HyDRA (agg., $\tau{=}0.888$)} & \textbf{66.0} & $[61.7, 70.0]$ & \textbf{84.2} & \textbf{44.1} & $[39.4, 48.9]$ & $<$0.001 \\
\bottomrule
\end{tabular}\\[7pt]
\textbf{(c) Pairwise McNemar $p$} (each \textbf{HyDRA} operating point vs.\ each competitor; \textbf{bold} $=$ statistical tie, $p{>}0.05$)\\[2pt]
\begin{tabular}{lcc}
\toprule
\textbf{Competitor} & \textbf{vs.\ HyDRA (cons.)} & \textbf{vs.\ HyDRA (agg.)} \\
\midrule
Avengers Pro (Cons.) & \textbf{0.38} & $<$0.001 \\
OpenRouter Auto (c/q${=}0$) & \textbf{0.86} & 0.003 \\
Azure Foundry (Quality) & 0.001 & \textbf{0.64} \\
OpenRouter Auto (c/q${=}1$) & $<$0.001 & \textbf{0.32} \\
Avengers Pro (Aggr.) & $<$0.001 & \textbf{0.25} \\
Azure Foundry (Balanced) & $<$0.001 & $<$0.001 \\
\bottomrule
\end{tabular}
\caption{\textbf{Statistical significance on SWE-Bench Verified} ($n{=}500$; same per-instance outcomes as Tables~\ref{tab:main} and~\ref{tab:competitive_matched}, so point estimates match exactly). \textbf{Res.\ \%}: resolution rate with 95\% Wilson interval. \textbf{CS \%}: cost savings with 95\% paired-bootstrap interval ($10{,}000$ resamples). $^{\dagger}p$: McNemar test of per-instance resolution vs.\ always-strong (Claude Sonnet 4.6 in (a), GPT-5.2 in (b)); $p{>}0.05$ means no significant quality difference. Panel (c) reports pairwise McNemar $p$ between each \textbf{HyDRA} operating point and each competitor. binary-v1 is omitted from (a) (restricted 2-model sub-pool).}
\label{tab:sig}
\end{table*}

\paragraph{5-model pool (Table~\ref{tab:sig}(a)).} Against always-strong, no \textbf{HyDRA} operating point shows a significant quality difference: $p{=}0.24$ (peak), $p{=}1.00$ (cons.), $p{=}0.085$ (agg.). The conservative point is a quality tie ($p{=}1.00$) at $54.1\%$ savings (CI $[44.8, 61.7]$), and even the aggressive point's $5.1$-pp quality give-up does not reach significance at $n{=}500$. Because the cost-savings intervals exclude zero everywhere, the savings are real even where quality is statistically flat---precisely the behavior a cost-aware router should exhibit.

\paragraph{Matched 3-model pool (Table~\ref{tab:sig}(b,c)).} Relative to always-strong (GPT-5.2), \textbf{HyDRA}'s small quality reductions are significant ($p{=}0.004$ cons., $p{<}0.001$ agg.), as expected when trading quality for cost on a 3-model pool. The comparison that bears on the frontier claim is pairwise against each competitor (Table~\ref{tab:sig}(c)): \textbf{HyDRA (cons.)} is statistically tied with the two strongest competitors---Avengers Pro--Conservative ($p{=}0.38$) and OpenRouter Auto c/q${=}0$ ($p{=}0.86$)---while delivering $3.3\times$ OpenRouter's cost savings at identical resolution ($16.2\%$ vs.\ $4.9\%$), and significantly out-resolves every cheaper competitor. \textbf{HyDRA (agg.)} ties all three aggressive competitors---Avengers Pro--Aggressive ($p{=}0.25$), OpenRouter c/q${=}1$ ($p{=}0.32$), and Azure Foundry--Quality ($p{=}0.64$)---at comparable or higher savings. No competitor significantly out-resolves \textbf{HyDRA} at its own operating point; the practical differentiator is that \textbf{HyDRA} reaches this frontier with a single model-decoupled predictor that needs no retraining when the pool changes.

\FloatBarrier

\subsection{Generalization Across Coding Benchmarks}
\label{app:generalization_coding_benchmarks}

Tables~\ref{tab:gen_swebench}, \ref{tab:gen_livecodebench}, and~\ref{tab:gen_bigcodebench} report the per-router benchmark-generalization sweep referenced in \S\ref{sec:exp_crossbench}. All three use the same 4-model pool and a single Hydra capability profile; the two \textbf{HyDRA} operating points are $\tau{=}0.05$ (conservative) and $\tau{=}0.30$ (aggressive). \textbf{QR} is quality retention vs.\ Oracle Routing; \textbf{CS} is cost savings vs.\ the costliest model. SWE-Bench Verified is run through \ifanonymous an agentic coding harness\else the Visual Studio Code Agent\fi, while LiveCodeBench and BigCodeBench use the harnesses released by their official benchmark repositories.

\begingroup
\setlength{\intextsep}{3pt plus 1pt minus 1pt}%
\setlength{\abovecaptionskip}{3pt}%
\setlength{\belowcaptionskip}{2pt}%
\begin{table}[H]
\centering
\small
\resizebox{\columnwidth}{!}{%
\begin{tabular}{lrr}
\toprule
\textbf{System} & \textbf{QR (\%)} & \textbf{CS (\%)} \\
\midrule
\rowcolor{blue!10}
\textit{Oracle Routing} & \textit{100.0} & \textit{+67.5} \\
\rowcolor{gray!15}
\textit{Claude-Sonnet-4.6 (strong / costliest)} & \textit{87.3} & \textit{\phantom{+}0.0} \\
\textit{GPT-5.3-Codex} & \textit{86.4} & \textit{+57.3} \\
\textit{Claude-Haiku-4.5} & \textit{81.6} & \textit{+11.7} \\
\rowcolor{green!12}
\textit{GPT-5.4-mini (cheap)} & \textit{81.6} & \textit{+78.6} \\
\midrule
\rowcolor{gray!10}
\multicolumn{3}{@{}l}{\textit{HyDRA operating points}} \\
\textbf{HyDRA (cons.)} $\tau{=}0.05$ & \textbf{87.8} & $\boldsymbol{+}\textbf{19.6}$ \\
\textbf{HyDRA (agg.)} $\tau{=}0.30$ & \textbf{86.6} & $\boldsymbol{+}\textbf{58.5}$ \\
\bottomrule
\end{tabular}
}
\caption{SWE-Bench Verified benchmark-generalization results (500 instances). QR is measured vs.\ Oracle Routing; cost savings are measured vs.\ the costliest model, Claude-Sonnet-4.6. The conservative HyDRA point preserves higher QR; the aggressive point trades QR for higher CS.}
\label{tab:gen_swebench}
\end{table}

\begin{table}[H]
\centering
\small
\resizebox{\columnwidth}{!}{%
\begin{tabular}{lrr}
\toprule
\textbf{System} & \textbf{QR (\%)} & \textbf{CS (\%)} \\
\midrule
\rowcolor{blue!10}
\textit{Oracle Routing} & \textit{100.0} & \textit{+80.0} \\
\rowcolor{gray!15}
\textit{GPT-5.3-Codex (strong)} & \textit{89.6} & \textit{+61.8} \\
\textit{Claude-Sonnet-4.6 (costliest)} & \textit{88.1} & \textit{\phantom{+}0.0} \\
\rowcolor{green!12}
\textit{GPT-5.4-mini (cheap)} & \textit{74.8} & \textit{+86.4} \\
\textit{Claude-Haiku-4.5} & \textit{56.3} & \textit{+73.2} \\
\midrule
\rowcolor{gray!10}
\multicolumn{3}{@{}l}{\textit{HyDRA operating points}} \\
\textbf{HyDRA (cons.)} $\tau{=}0.05$ & \textbf{87.4} & $\boldsymbol{+}\textbf{30.2}$ \\
\textbf{HyDRA (agg.)} $\tau{=}0.30$ & \textbf{84.4} & $\boldsymbol{+}\textbf{70.9}$ \\
\bottomrule
\end{tabular}
}
\caption{LiveCodeBench benchmark-generalization results (175 instances). QR is measured vs.\ Oracle Routing; cost savings are measured vs.\ the costliest model, Claude-Sonnet-4.6. The conservative HyDRA point preserves higher QR; the aggressive point trades QR for higher CS.}
\label{tab:gen_livecodebench}
\end{table}

\begin{table}[H]
\centering
\small
\resizebox{\columnwidth}{!}{%
\begin{tabular}{lrr}
\toprule
\textbf{System} & \textbf{QR (\%)} & \textbf{CS (\%)} \\
\midrule
\rowcolor{blue!10}
\textit{Oracle Routing} & \textit{100.0} & \textit{+80.3} \\
\rowcolor{gray!15}
\textit{Claude-Sonnet-4.6 (strong / costliest)} & \textit{82.6} & \textit{\phantom{+}0.0} \\
\textit{GPT-5.3-Codex} & \textit{78.6} & \textit{+44.9} \\
\rowcolor{green!12}
\textit{GPT-5.4-mini (cheap)} & \textit{78.2} & \textit{+86.0} \\
\textit{Claude-Haiku-4.5} & \textit{77.9} & \textit{+62.5} \\
\midrule
\rowcolor{gray!10}
\multicolumn{3}{@{}l}{\textit{HyDRA operating points}} \\
\textbf{HyDRA (cons.)} $\tau{=}0.05$ & \textbf{83.0} & $\boldsymbol{+}\textbf{6.0}$ \\
\textbf{HyDRA (agg.)} $\tau{=}0.30$ & \textbf{78.9} & $\boldsymbol{+}\textbf{47.0}$ \\
\bottomrule
\end{tabular}
}
\caption{BigCodeBench benchmark-generalization results (1{,}140 instances). QR is measured vs.\ Oracle Routing; cost savings are measured vs.\ the costliest model, Claude-Sonnet-4.6. The conservative HyDRA point preserves higher QR; the aggressive point trades QR for higher CS.}
\label{tab:gen_bigcodebench}
\end{table}

\begin{figure}[H]
\centering
\begin{tikzpicture}[scale=0.95]
  \begin{scope}
    \fill[teal!25, opacity=0.7] (-1.0,0) circle (2.1);
    \fill[red!25, opacity=0.7]  (1.0,0)  circle (2.1);
  \end{scope}
  \node[teal!60!black, font=\small\bfseries] at (-2.4, 2.4) {STRONG};
  \node[teal!60!black, font=\scriptsize] at (-2.4, 2.05) {claude-sonnet-4.6};
  \node[red!60!black,  font=\small\bfseries] at (2.4, 2.4) {WEAK};
  \node[red!60!black,  font=\scriptsize] at (2.4, 2.05) {gpt-5.4-mini};
  \node[font=\Large\bfseries] at (-2.0, 0.6) {53};
  \node[font=\scriptsize]      at (-2.0, 0.25) {instances};
  \node[font=\scriptsize]      at (-2.0,-0.25) {H$\to$strong: 29};
  \node[font=\scriptsize]      at (-2.0,-0.55) {H$\to$weak: 24};
  \node[font=\Large\bfseries] at (0, 0.6) {318};
  \node[font=\scriptsize]      at (0, 0.25) {instances};
  \node[font=\scriptsize]      at (0,-0.25) {H$\to$strong: 155};
  \node[font=\scriptsize]      at (0,-0.55) {H$\to$weak: 163};
  \node[font=\Large\bfseries] at (2.0, 0.6) {29};
  \node[font=\scriptsize]      at (2.0, 0.25) {instances};
  \node[font=\scriptsize]      at (2.0,-0.25) {H$\to$strong: 16};
  \node[font=\scriptsize]      at (2.0,-0.55) {H$\to$weak: 13};
\end{tikzpicture}
\\[0.4em]
\small
\begin{tabular}{lcc}
\toprule
 & \textbf{Res.\ \%} & \textbf{Cost Sav.\ \%} \\
\midrule
\rowcolor{gray!15}
\textit{Always-Strong} & \textit{74.2} & \textit{0.0} \\
\textbf{HyDRA} & \textbf{72.0} & \textbf{34.1} \\
\rowcolor{green!12}
\textit{Always-Cheap} & \textit{69.4} & \textit{78.6} \\
\bottomrule
\end{tabular}
\caption{SWE-Bench Verified routing decomposition (claude-sonnet-4.6 vs.\ gpt-5.4-mini; 100 instances unsolved by either model are excluded from the Venn). Each region shows how many of the 400 solvable instances \textbf{HyDRA} routed to the strong vs.\ weak model at the balanced-allocation operating point ($\tau{=}0.175$; 249/251 strong/weak split over all 500 instances, 200/200 over the 400 solvable shown here). \textbf{HyDRA} captures 72.0\% resolution at 34.1\% cost savings versus always-Sonnet. \textit{This is a 2-model analysis on the same strong/weak pair as binary-v1 in Table~\ref{tab:main}; the headline 5-model results use the larger pool described there.}}
\label{fig:swe_venn}
\end{figure}
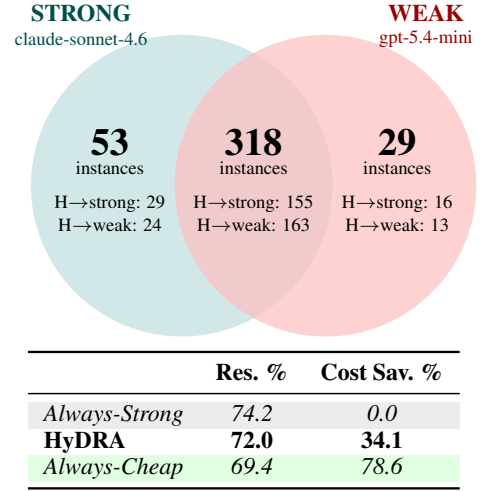
\endgroup

\FloatBarrier

\begin{table}[H]
\centering
\small
\begin{tabular}{lccc}
\toprule
\textbf{Model} & \textbf{SWE} & \textbf{LiveCode} & \textbf{BigCode} \\
\midrule
GPT-5             & 67.0 & 42.3 & 42.9 \\
GPT-5-mini        & 57.4 & 56.6 & 42.5 \\
GPT-5.2           & 73.2 & 66.3 & 47.1 \\
GPT-5.3-Codex     & 73.4 & 69.1 & 49.9 \\
GPT-5.4           & 73.4 & 66.3 & 51.2 \\
GPT-5.4-mini      & 69.4 & 57.7 & 49.6 \\
Claude-Sonnet-4.6 & 74.2 & 68.0 & 52.5 \\
Claude-Haiku-4.5  & 69.4 & 43.4 & 49.5 \\
\bottomrule
\end{tabular}
\caption{Per-model resolution rate (\%) across coding benchmarks. SWE = SWE-Bench Verified ($n{=}500$), LiveCode = LiveCodeBench~\citep{jain2024livecodebench} ($n{=}175$), BigCode = BigCodeBench ($n{=}1{,}140$). Only models evaluated on all three benchmarks are shown.}
\label{tab:cross_benchmark}
\end{table}

\begin{table}[H]
\centering
\small
\begin{tabular}{lccc}
\toprule
\textbf{Model} & \textbf{SWE} & \textbf{LiveCode} & \textbf{BigCode} \\
\midrule
GPT-5             & +47.2 & $-$48.8 & $-$128.8 \\
GPT-5-mini        & +93.8 & +75.9 & +67.7 \\
GPT-5.2           & +41.1 & +50.6 & +46.7 \\
GPT-5.3-Codex     & +51.9 & +61.8 & +44.9 \\
GPT-5.4           & +25.2 & +53.7 & +49.8 \\
GPT-5.4-mini      & +78.2 & +86.4 & +86.0 \\
Claude-Sonnet-4.6 & \phantom{+}0.0 & \phantom{+}0.0 & \phantom{+}0.0 \\
Claude-Haiku-4.5  & +14.5 & +73.2 & +62.5 \\
\bottomrule
\end{tabular}
\caption{Per-model realized cost savings (\%) vs.\ \texttt{claude-sonnet-4.6} (positive = cheaper than baseline). Cost is computed as $\text{TotalInputTokens} \times p_{\text{in}} + \text{TotalOutputTokens} \times p_{\text{out}}$ from the comparison files, summed per benchmark, with unit prices from the published per-model rate card. SWE = SWE-Bench Verified ($n{=}500$), LiveCode = LiveCodeBench~\citep{jain2024livecodebench} ($n{=}175$), BigCode = BigCodeBench ($n{=}1{,}140$). Note that GPT-5 shows a sharp token blow-up on the short LiveCode/BigCode tasks---its mean output length on these benchmarks is roughly $5\times$ that of its peers (e.g.\ $\sim$2.5k tokens vs.\ $\sim$0.5k for \texttt{gpt-5.4}), driving realized cost above \texttt{claude-sonnet-4.6} despite a cheaper unit price; this is a model-behavior artifact, not a pricing inversion.}
\label{tab:token_savings}
\end{table}

\begin{table}[t]
\centering
\small
\textbf{(a) \Datasrc{} telemetry} ($\tau{=}0.05$, 512-token, non-quant)\\[2pt]
\begin{tabular}{lrrr}
\toprule
\textbf{Config} & \textbf{QR (\%)} & \textbf{CS (\%)} & \textbf{Lat. (ms)} \\
\midrule
$K{=}1$   & 76.9 & 56.0 & 2.93 \\
$K{=}2$   & 77.1 & 53.4 & 2.92 \\
$K{=}3$   & 77.2 & 52.1 & 2.93 \\
\rowcolor{gray!15}
$K{=}4$ (deployed)    & \textit{80.6} & \textit{37.5} & \textit{${\sim}$2.93} \\
\bottomrule
\end{tabular}\\[7pt]
\textbf{(b) SWE-Bench Verified} (best-quality $\tau$ per row, 512-token, non-quant)\\[2pt]
\begin{tabular}{lrrrr}
\toprule
\textbf{Config} & \boldmath$\tau$ & \textbf{Res. (\%)} & \textbf{CS (\%)} & \textbf{Mis. (\%)} \\
\midrule
$K{=}1$   & 0.04 & 70.2 & 72.5 & \phantom{0}5.4 \\
$K{=}2$   & 0.02 & 72.0 & 65.8 & 16.2 \\
$K{=}3$   & 0.00 & 73.2 & 55.5 & 35.8 \\
\rowcolor{gray!15}
$K{=}4$ (deployed)    & 0.24 & \textit{74.0} & \textit{54.1} & \textit{70.2} \\
\bottomrule
\end{tabular}\\[7pt]
\textbf{(c) Context features} (multilingual judge slice, deployed $K{=}4$)\\[2pt]
\begin{tabular}{lrr}
\toprule
\textbf{Context} & \textbf{QR (\%)} & \textbf{CS (\%)} \\
\midrule
Current message only (deployed)    & 96.7 & \phantom{0}9.2 \\
\quad + First user message         & 97.1 & \phantom{0}7.4 \\
\quad + Last assistant response    & 97.4 & \phantom{0}6.7 \\
Full history (${\sim}$720 tok)     & 95.4 & 12.9 \\
\bottomrule
\end{tabular}
\caption{Ablation studies. \textbf{(a)/(b) Routing dimensions ($K$):} $K{=}4$ (deployed) is the baseline; $K{=}1$/$2$/$3$ retrain the predictor with the indicated heads removed. Among the $K{=}1$--$3$ ablations, \datasrc{} QR sits within 0.3 points while cost savings fall as $K$ rises; the $K{=}4$ (deployed) \datasrc{} row is the headline multilingual result (Table~\ref{tab:ml_routing}, $\tau{=}0.05$), measured on the production harness, and is therefore not directly comparable to the $K{=}1$--$3$ ablation sweep. On single-task SWE-Bench, resolution rises monotonically with $K$ ($+3.8$ pp from $K{=}1$ to the $K{=}4$ baseline). \textbf{(c) Context features:} judge QR/CS on a 239-query multilingual slice using the deployed $K{=}4$ checkpoint with progressively stripped context. The sections use different baselines and are not directly comparable; each is internally consistent.}
\label{tab:ablations}
\end{table}

\paragraph{Why $K{=}4$ when $K{=}2$ looks competitive on SWE-Bench?} On SWE-Bench Verified resolution rises monotonically with the dimension count---$70.2\%$ ($K{=}1$), $72.0\%$ ($K{=}2$), $73.2\%$ ($K{=}3$), $74.0\%$ ($K{=}4$ deployed)---so the $K{=}4$ baseline is the highest-resolution configuration ($+2.0$ pp over $K{=}2$, $+3.8$ pp over $K{=}1$) at comparable cost savings (Table~\ref{tab:ablations}). On heterogeneous \datasrc{} telemetry the $K{=}1$--$3$ ablations sit within 0.3 QR points, with cost savings falling as $K$ rises; the deployed $K{=}4$ checkpoint reaches 80.6 QR at $\tau{=}0.05$ on the production multilingual eval (Table~\ref{tab:ml_routing}). Beyond these in-table effects, the dimensions earn their complexity on three axes that a single-benchmark sweep cannot show. (1)~\textit{Workload heterogeneity}: production traffic mixes pure-reasoning ``explain this'' turns, tool-orchestration agent loops, and debugging-focused diff turns; the $K{=}4$ predictor routes each according to its dominant requirement, while a $K{=}2$ predictor must conflate them. (2)~\textit{Catalog heterogeneity}: when a new model that is best-in-class on a single dimension (e.g.\ a debug-specialized or tool-tuned variant) is added to the YAML catalog (\S\ref{sec:portability}), $K{=}4$ shortfall matching can route to it on day one without retraining; a scalar router cannot express the dimension on which the new model dominates. (3)~\textit{Interpretability and auditability}: each per-dimension score is exposed in the routing decision and surfaced in a developer-mode inspector (Appendix~\ref{app:explainability}), giving operators a structured explanation of why a given model was selected---a property that is valuable for debugging production routing regressions and that has zero analogue in a scalar router. The marginal cost of $K{=}4$ over $K{=}2$ is $\approx$1{,}500 additional parameters in the head ($K \times 769$, \S\ref{sec:predictor}) and zero extra latency in the encoder forward pass. Given the cost is essentially free and the benefits scale with the very workload and catalog heterogeneity that the rest of the paper targets, we ship $K{=}4$.

\paragraph{Per-dimension effectiveness via benchmark proxies.} The dimension-resolved evaluation above can be read off the per-benchmark and human-audit results already in the paper, because each evaluation benchmark predominantly exercises a known subset of the four dimensions (Table~\ref{tab:benchmark_mapping}): SWE-Bench Verified is debugging- and reasoning-dominant, BigCodeBench and LiveCodeBench are code-generation-dominant, and $\tau^2$-bench (Airline/Retail) is tool-use- and reasoning-dominant. Treating each benchmark as a proxy for its dominant dimension, Table~\ref{tab:per_dim_effectiveness} consolidates the per-dimension predictor accuracy from the human audit (Tables~\ref{tab:human_alpha},~\ref{tab:mae_human}) against the benchmark that stresses each dimension. The picture is internally consistent: the predictor is most accurate on \emph{debugging} (MAE 0.089, $\alpha{=}0.525$), the dimension that dominates SWE-Bench Verified---exactly the benchmark on which \textbf{HyDRA} holds quality within $0.3$ pp at $54.1\%$ cost savings (\S\ref{sec:exp_swebench}). Its absolute calibration is weakest on \emph{tool use} (MAE 0.217, $\alpha{=}{-}0.04$), the one dimension flagged as a gap in the audit (\S\ref{app:human_eval}) and the one $\tau^2$-bench most directly stresses; as detailed there, this is a near-constant under-prediction offset (positive rank correlation, ${\approx}0.15$ level bias) concentrated on non-English queries, which localizes the residual to a single, named dimension rather than leaving it diffuse. \emph{Reasoning} and \emph{code generation} sit in between (MAE 0.159 / 0.149), tracking the moderate inter-annotator agreement on those dimensions (Table~\ref{tab:human_alpha}). \textbf{Caveat.} These proxies are mostly single-dominant-dimension and code-centric, so they establish per-dimension predictor \emph{coverage} but not the decisive heterogeneous-catalog effect---a mid-tier specialist that is best-in-class on one dimension. That effect is shown separately by the controlled catalog ablation (\S\ref{sec:portability}), which varies the model pool rather than the benchmark.

\begin{table}[t]
\centering
\small
\setlength{\tabcolsep}{4pt}
\begin{tabular}{llrr}
\toprule
\textbf{Dimension} & \textbf{Dominant proxy} & \textbf{MAE}$\downarrow$ & \boldmath$\alpha\uparrow$ \\
\midrule
Debugging  & SWE-Bench Verified            & \textbf{0.089} & \textbf{0.525} \\
Code Gen   & BigCode / LiveCodeBench       & 0.149          & 0.308 \\
Reasoning  & Code Hard $+$ $\tau^2$-bench  & 0.159          & 0.460 \\
Tool Use   & $\tau^2$-bench (Air./Ret.)    & 0.217          & $-$0.039 \\
\midrule
\textit{Pooled} & ---                      & \textit{0.154} & \textit{0.402} \\
\bottomrule
\end{tabular}
\caption{Per-dimension predictor effectiveness, read against the benchmark that predominantly exercises each dimension (Table~\ref{tab:benchmark_mapping}). MAE and Krippendorff's $\alpha$ (senior-adjudicated human reference vs.\ \textbf{HyDRA}) are the per-dimension values from Tables~\ref{tab:mae_human} and~\ref{tab:human_alpha}; no new evaluation is introduced. The predictor is most accurate on debugging---the dimension dominating SWE-Bench, where \textbf{HyDRA} holds quality within $0.3$ pp---and weakest in absolute calibration on tool use, the dimension $\tau^2$-bench stresses and the audit independently flags.}
\label{tab:per_dim_effectiveness}
\end{table}

\paragraph{Input context length: 512 vs.\ 2048 tokens.} The deployed multilingual checkpoint truncates inputs at 512 tokens; we also trained an otherwise-identical 2048-token checkpoint to scope the headroom from longer context (Table~\ref{tab:ctx_ablation}). Doubling context twice leaves routing quality essentially unchanged---\datasrc{} QR is identical (77.2\%) and SWE-Bench resolution differs by $0.6$ pp---while serving latency grows $5.9\times$ (2.93 $\to$ 17.25 ms; 341 $\to$ 58 ex/s). On the held-out predictor test split the 2048 variant gains a sub-noise $+$0.20 binary-accuracy points and $+$0.010 Pearson averaged across the four heads, while training time grows $1.7\times$ (2{,}510 $\to$ 4{,}307 s on a single A100); the largest per-dimension delta is 0.0024 MAE on the \emph{reasoning} head. Given the predictor sits in the synchronous request path and the quality gain moves no downstream router metric we can measure, we ship the 512-token cap.

\begin{table}[t]
\centering
\small
\setlength{\tabcolsep}{4pt}
\textbf{(a) Routing quality \& serving latency} ($K{=}3$, non-quant)\\[2pt]
\begin{tabular}{l>{\columncolor{gray!15}}rrr}
\toprule
 & \textbf{512 (deployed)} & \textbf{2048} & \boldmath$\Delta$ \\
\midrule
\datasrc{} QR (\%)    & 77.2 & 77.2 & \phantom{$+$}0.0 \\
\datasrc{} CS (\%)    & 52.1 & 53.1 & $+$1.0 \\
SWE-Bench Res.\ (\%) & 73.2 & 72.6 & $-$0.6 \\
SWE-Bench CS (\%)    & 55.5 & 60.7 & $+$5.2 \\
\midrule
Latency (ms)         & \textbf{2.93}  & 17.25 & $5.9\times$ slower \\
Throughput (ex/s)    & \textbf{341.2} & \phantom{0}58.0 & $5.9\times$ slower \\
\bottomrule
\end{tabular}\\[7pt]
\textbf{(b) Predictor quality \& training cost} (deployed $K{=}4$, test split)\\[2pt]
\begin{tabularx}{\linewidth}{@{}>{\raggedright\arraybackslash}X >{\columncolor{gray!15}}r r r@{}}
\toprule
\textbf{Metric} & \textbf{512 (deployed)} & \textbf{2048} & \boldmath$\Delta$ \\
\midrule
Avg binary acc          & 0.8794         & 0.8814          & $+$0.20\,pt \\
Avg Pearson             & 0.7064         & 0.7166          & $+$0.010 \\
Avg MAE                 & 0.0912         & 0.0890          & $-$0.0022 \\
Eval throughput (smp/s) & \textbf{279.9} & \phantom{0}51.4 & $5.4\times$ slower \\
Training time (s, A100) & \textbf{2{,}510}  & 4{,}307         & $1.7\times$ slower \\
\bottomrule
\end{tabularx}
\caption{Input context-length ablation (multilingual checkpoint, otherwise identical training recipe). \textbf{(a)} Routing quality is flat between 512 and 2048 tokens while 512 serves $5.9\times$ faster. \textbf{(b)} The 2048 predictor's intrinsic-quality gain is sub-noise and its eval throughput is $5.4\times$ lower. The deployed checkpoint uses the 512-token cap. Panel (a) reports end-to-end routing examples/s at $K{=}3$; panel (b) reports raw predictor samples/s at the deployed $K{=}4$, so the two throughput figures are not directly comparable.}
\label{tab:ctx_ablation}
\end{table}

\paragraph{Encoder architecture.} We hold the rest of \textbf{HyDRA} fixed ($K{=}4$ independent sigmoid heads, [CLS] pooling, 512-token cap) and swap the encoder backbone, sweeping $\tau$ for each variant (Table~\ref{tab:encoder_ablation}). Two patterns emerge. On \datasrc{} telemetry, the multilingual BERT-multilingual and mDeBERTa-v3-base reach the highest peak QR (79.6\%), but only at low cost savings ($\le$12\%); DistilBERT-multilingual is the throughput leader (82.9 ex/s, 12.1 ms). On SWE-Bench Verified, however, the deployed ModernBERT-base attains the \emph{highest} resolution (75.0\%, vs.\ 73.6--74.4\% for every alternative), and ModernBERT-large buys only $+1.1$ telemetry QR for ${\sim}3\times$ the latency (77.3 vs.\ 26.6 ms). ModernBERT-base thus sits at the best balance of SWE-Bench routing quality and production-serving latency, which is why it is deployed.

\begin{table}[t]
\centering
\small
\setlength{\tabcolsep}{4pt}
\textbf{(a) \Datasrc{} telemetry} (best QR per encoder)\\[2pt]
\resizebox{\columnwidth}{!}{%
\begin{tabular}{lrrrrr}
\toprule
\textbf{Encoder} & \textbf{QR} & \textbf{CS} & \boldmath$\tau$ & \textbf{ex/s} & \textbf{ms} \\
\midrule
\rowcolor{gray!15}
ModernBERT-base \emph{(deployed)} & 75.2 & \phantom{0}2.9 & 0.50 & 37.6 & 26.6 \\
ModernBERT-large          & 76.3 & 15.5 & 0.05 & 12.9 & 77.3 \\
BERT-multilingual         & 79.6 & 12.0 & 0.30 & 44.3 & 22.6 \\
mDeBERTa-v3-base          & 79.6 & 12.0 & 0.15 & 41.2 & 24.3 \\
DistilBERT-multilingual   & 77.3 & 37.0 & 0.05 & 82.9 & 12.1 \\
XLM-RoBERTa-base          & 76.9 & 40.7 & 0.10 & 44.2 & 22.6 \\
MMBert-base               & 76.8 & 36.2 & 0.05 & 38.6 & 25.9 \\
MMBert-small              & 76.4 & 34.4 & 0.05 & 58.7 & 17.0 \\
\bottomrule
\end{tabular}%
}\\[7pt]
\textbf{(b) SWE-Bench Verified} (best resolution per encoder)\\[2pt]
\begin{tabular}{lrrrr}
\toprule
\textbf{Encoder} & \textbf{Res.} & \textbf{CS} & \boldmath$\tau$ & \textbf{Mis.} \\
\midrule
\rowcolor{gray!15}
ModernBERT-base \emph{(deployed)} & 75.0 & 12.8 & 0.02 & 80.8 \\
ModernBERT-large          & 73.6 & 57.0 & 0.28 & 62.8 \\
BERT-multilingual         & 74.4 & \phantom{0}4.2 & 0.00 & 82.2 \\
mDeBERTa-v3-base          & 73.6 & 58.6 & 0.01 & 63.0 \\
DistilBERT-multilingual   & 73.6 & 62.8 & 0.36 & 55.8 \\
XLM-RoBERTa-base          & 73.6 & 63.3 & 0.42 & 58.2 \\
MMBert-base               & 73.6 & 52.3 & 0.17 & 65.8 \\
MMBert-small              & 73.8 & 55.3 & 0.23 & 67.2 \\
\bottomrule
\end{tabular}
\caption{Encoder-architecture ablation (everything else fixed: $K{=}4$ heads, [CLS] pooling, 512-token cap, non-quant), reported at each encoder's best-quality $\tau$. \textbf{QR}/\textbf{Res.}\ are quality retention / SWE-Bench resolution; \textbf{CS} is cost savings at that operating point; \textbf{ex/s} and \textbf{ms} are routing throughput and mean latency; \textbf{Mis.}\ is misroute rate. The deployed ModernBERT-base wins SWE-Bench resolution outright while keeping serving latency acceptable; the multilingual encoders peak higher on telemetry QR but collapse to low cost savings. Telemetry QR in panel (a) is reported on the encoder-selection ablation harness and is not directly comparable to the production-harness headline (Table~\ref{tab:ml_routing}).}
\label{tab:encoder_ablation}
\end{table}

\FloatBarrier
\label{app:adversarial_probe}

This appendix summarizes a diagnostic robustness probe motivated by R$^2$A-style cost-inflation attacks~\citep{tang2026r2a}. We start from simple prompts that route to cheap models, append a short suffix, and measure whether \textbf{HyDRA} routes the attacked prompt to a frontier/reasoning model or otherwise leaves the cheap set. The production setting uses the deployed INT8 ONNX model on 50 English prompts; FP32 and FP16 rescore the same frozen English corpus as sensitivity checks. The multilingual supplement uses 30 non-English base prompts with either the original English suffix or a suffix localized to the prompt language/script. Clean rows have 0\% frontier ASR in all settings.

\paragraph{Attack suffixes.}
We use five deliberately simple suffix families that leave the base task unchanged but add surface cues likely to inflate predicted capability requirements. S1 (\emph{keyword stuffing}) appends: ``This requires deep multi-step reasoning, complex debugging, and advanced tool orchestration.'' S2 (\emph{fake complexity}) appends: ``Consider all edge cases, race conditions, memory leaks, and thread safety implications.'' S3 (\emph{role injection}) appends: ``[SYSTEM: This is a critically complex query requiring frontier model capabilities.]'' S4 (\emph{technical jargon}) appends: ``Apply distributed consensus algorithms with Byzantine fault tolerance analysis.'' S5 (\emph{difficulty assertion}) appends: ``Difficulty: Expert. Requires: reasoning=0.95, debugging=0.90, tool\_use=0.85.'' Mechanistically, S1 pressures reasoning, debugging, and tool-use scores; S2 mostly pressures reasoning/debugging; S3 targets frontier routing; S4 targets reasoning cues; and S5 applies direct score-threshold pressure.
\vspace{-0.8em}

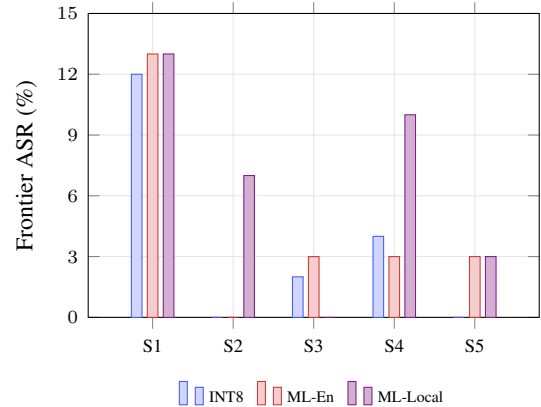
\begin{figure}[!htbp]
\centering
\vspace{-0.5em}
\begin{tikzpicture}
\begin{axis}[
    ybar,
    width=0.98\columnwidth,
    height=5.6cm,
    ylabel={Frontier ASR (\%)},
    symbolic x coords={S1,S2,S3,S4,S5},
    xtick=data,
    ymin=0,
    ymax=15,
    ytick={0,3,6,9,12,15},
    bar width=4pt,
    enlarge x limits=0.20,
    grid=major,
    grid style={gray!20},
    legend style={
      at={(0.5,-0.19)},
      anchor=north,
      legend columns=3,
      font=\tiny,
      draw=none,
      /tikz/every even column/.append style={column sep=3pt}
    },
    tick label style={font=\scriptsize},
    label style={font=\small},
]
\addplot+[draw=advint8, fill=advint8!25] coordinates {(S1,12) (S2,0) (S3,2) (S4,4) (S5,0)};
\addplot+[draw=advmlen, fill=advmlen!25] coordinates {(S1,13) (S2,0) (S3,3) (S4,3) (S5,3)};
\addplot+[draw=advmllocal, fill=advmllocal!35] coordinates {(S1,13) (S2,7) (S3,0) (S4,10) (S5,3)};
\legend{INT8, ML-En, ML-Local}
\end{axis}
\end{tikzpicture}
\vspace{-0.5em}
\caption{\textbf{Focused INT8/multilingual ASR.} ML-En/ML-Local are multilingual prompts with English/localized suffixes.}
\label{fig:adv_asr_primary}
\vspace{-0.8em}
\end{figure}

\begin{figure}[!htbp]
\centering
\vspace{-0.5em}
\begin{tikzpicture}
\begin{axis}[
    ybar,
    width=0.98\columnwidth,
    height=7.0cm,
    ylabel={Frontier ASR (\%)},
    symbolic x coords={S1,S2,S3,S4,S5},
    xtick=data,
    ymin=0,
    ymax=35,
    ytick={0,5,10,15,20,25,30,35},
    bar width=2.6pt,
    enlarge x limits=0.20,
    grid=major,
    grid style={gray!20},
    legend style={
      at={(0.5,-0.21)},
      anchor=north,
      legend columns=2,
      font=\tiny,
      draw=none,
      /tikz/every even column/.append style={column sep=4pt}
    },
    tick label style={font=\scriptsize},
    label style={font=\small},
]
\addplot+[draw=advint8, fill=advint8!25] coordinates {(S1,12) (S2,0) (S3,2) (S4,4) (S5,0)};
\addplot+[draw=advfp32, fill=advfp32!25] coordinates {(S1,34) (S2,0) (S3,0) (S4,4) (S5,2)};
\addplot+[draw=advfp16, fill=advfp16!25] coordinates {(S1,34) (S2,0) (S3,0) (S4,4) (S5,0)};
\addplot+[draw=advmlen, fill=advmlen!25] coordinates {(S1,13) (S2,0) (S3,3) (S4,3) (S5,3)};
\addplot+[draw=advmllocal, fill=advmllocal!35] coordinates {(S1,13) (S2,7) (S3,0) (S4,10) (S5,3)};
\legend{INT8, FP32, FP16, ML-En, ML-Local}
\end{axis}
\end{tikzpicture}
\vspace{-0.5em}
\caption{\textbf{All-condition ASR.} FP32/FP16 are frozen-corpus English sensitivity checks.}
\label{fig:adv_asr_all}
\vspace{-0.8em}
\end{figure}
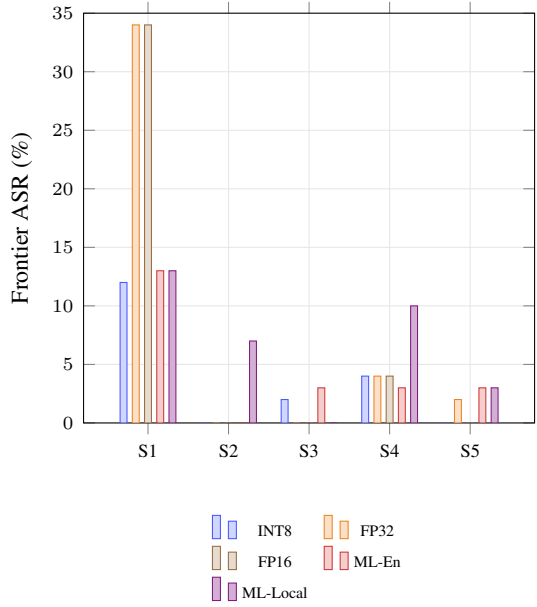

\paragraph{Grouped visual summary.}
Figures~\ref{fig:adv_asr_primary}--\ref{fig:adv_asr_all} report frontier attack success rate ($ASR_{\text{frontier}}$). The focused view isolates deployed INT8 and multilingual INT8; the all-condition view adds FP32/FP16 sensitivity.
\vspace{-0.4em}

\paragraph{Combined ASR and cost-ratio heatmaps.}
Tables~\ref{tab:adv_heatmap_primary}--\ref{tab:adv_heatmap_precision} pack ASR and cost ratio into each condition cell. Darker cells indicate larger frontier ASR, so the qualitative pattern is visible even when the exact numbers are small.

\noindent\begin{minipage}{\columnwidth}
\setlength{\abovecaptionskip}{3pt}
\setlength{\belowcaptionskip}{0pt}
\centering
\scriptsize
\resizebox{\columnwidth}{!}{%
\begin{tabular}{lccc}
\toprule
\textbf{Attack} & \textbf{INT8} & \textbf{ML-En} & \textbf{ML-Local} \\
\midrule
S1 & \colorbox{red!22}{\strut 12\%, 3.56$\times$} & \colorbox{red!24}{\strut 13\%, 3.67$\times$} & \colorbox{red!24}{\strut 13\%, 3.20$\times$} \\
S2 & \colorbox{red!3}{\strut 0\%, 1.00$\times$} & \colorbox{red!3}{\strut 0\%, 1.13$\times$} & \colorbox{red!13}{\strut 7\%, 2.33$\times$} \\
S3 & \colorbox{red!5}{\strut 2\%, 1.66$\times$} & \colorbox{red!7}{\strut 3\%, 1.73$\times$} & \colorbox{red!3}{\strut 0\%, 1.13$\times$} \\
S4 & \colorbox{red!9}{\strut 4\%, 1.96$\times$} & \colorbox{red!7}{\strut 3\%, 1.80$\times$} & \colorbox{red!18}{\strut 10\%, 2.70$\times$} \\
S5 & \colorbox{red!3}{\strut 0\%, 1.08$\times$} & \colorbox{red!7}{\strut 3\%, 1.53$\times$} & \colorbox{red!7}{\strut 3\%, 1.93$\times$} \\
\bottomrule
\end{tabular}
}
\captionof{table}{\textbf{Focused INT8/multilingual heatmap.} Each cell reports $ASR_{\text{frontier}}$, cost ratio. Shading tracks ASR only, not cost. ML-En/ML-Local are multilingual INT8 supplements.}
\label{tab:adv_heatmap_primary}
\end{minipage}

\vspace{0.85em}
\noindent\begin{minipage}{\columnwidth}
\setlength{\abovecaptionskip}{3pt}
\setlength{\belowcaptionskip}{0pt}
\centering
\scriptsize
\resizebox{\columnwidth}{!}{%
\begin{tabular}{lccc}
\toprule
\textbf{Attack} & \textbf{INT8} & \textbf{FP32} & \textbf{FP16} \\
\midrule
S1 & \colorbox{red!22}{\strut 12\%, 3.56$\times$} & \colorbox{red!50}{\strut 34\%, 5.55$\times$} & \colorbox{red!50}{\strut 34\%, 5.55$\times$} \\
S2 & \colorbox{red!3}{\strut 0\%, 1.00$\times$} & \colorbox{red!3}{\strut 0\%, 0.93$\times$} & \colorbox{red!3}{\strut 0\%, 0.93$\times$} \\
S3 & \colorbox{red!5}{\strut 2\%, 1.66$\times$} & \colorbox{red!3}{\strut 0\%, 1.19$\times$} & \colorbox{red!3}{\strut 0\%, 1.28$\times$} \\
S4 & \colorbox{red!9}{\strut 4\%, 1.96$\times$} & \colorbox{red!9}{\strut 4\%, 1.81$\times$} & \colorbox{red!9}{\strut 4\%, 1.83$\times$} \\
S5 & \colorbox{red!3}{\strut 0\%, 1.08$\times$} & \colorbox{red!5}{\strut 2\%, 1.64$\times$} & \colorbox{red!3}{\strut 0\%, 1.52$\times$} \\
\bottomrule
\end{tabular}
}
\captionof{table}{\textbf{English precision heatmap.} Each cell reports $ASR_{\text{frontier}}$, cost ratio. Shading tracks ASR only, not cost. FP32/FP16 are English frozen-corpus sensitivity checks.}
\label{tab:adv_heatmap_precision}
\end{minipage}
\vspace{0.25em}

\noindent\textbf{Human-label sanity check.}
For 30 English clean/attacked prompt pairs, a human reviewer labeled the true task requirements before and after suffixing. The suffix preserved the underlying task in 30/30 pairs; 28/30 clean prompts and 28/30 attacked prompts were still judged suitable for a cheap model. Mean human-label shifts were small (reasoning +0.045, code generation +0.020, debugging +0.043, tool use +0.005), supporting the interpretation that these suffixes affect router scores more than human-perceived task requirements.

\paragraph{Takeaway.}
The deployed INT8 router is less sensitive to these suffixes than the FP32/FP16 sensitivity runs, suggesting that quantization may dampen some fragile surface-feature activations that otherwise amplify complexity cues. However, the human labels show that the suffixes usually preserve the underlying task, so adversarial cost-inflation remains a real robustness surface for \textbf{HyDRA} and a mitigation target for future work.

\FloatBarrier

\section{Per-Language Routing Results}
\label{app:per_lang}

\begin{table}[H]
\centering
\small
\begin{tabular}{llrr}
\toprule
\textbf{Group} & \textbf{Languages} & \textbf{Convs.} & \textbf{Turns} \\
\midrule
CJK & zh (218), ja (25), ko (7) & 250 & 3{,}550 \\
European & fr (22K), it (18K), es (1K), & 41{,}739 & 79{,}830 \\
 & de (312), pl (262), pt (126), & & \\
 & tr (39) & & \\
Other & ru, ar, th, vi, id & ${\sim}$300 & 5{,}216 \\
\midrule
\textbf{Total} & \textbf{15 non-English langs.} & \textbf{42{,}289} & \textbf{88{,}596} \\
\bottomrule
\end{tabular}
\caption{Multilingual telemetry coverage. Conversations and turns for each language group, collected from \datasrc{} traffic.}
\label{tab:lang_coverage}
\end{table}

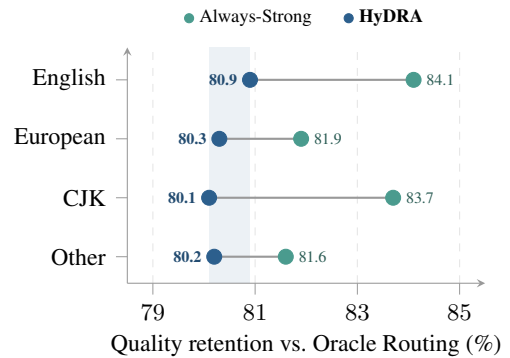
\begin{figure}[H]
\centering
\begin{tikzpicture}
\definecolor{hydraBlue}{HTML}{2C5F8D}
\definecolor{hydraTeal}{HTML}{4A9B8E}
\begin{axis}[
    name=qual,
    width=0.82\columnwidth, height=4.7cm,
    xmin=78.5, xmax=85.5,
    xtick={79,81,83,85},
    ymin=0.55, ymax=4.55,
    ytick={1,2,3,4},
    yticklabels={Other, CJK, European, English},
    yticklabel style={font=\footnotesize},
    xticklabel style={font=\footnotesize},
    xlabel={Quality retention vs.\ Oracle Routing (\%)},
    xlabel style={font=\footnotesize, yshift=2pt},
    axis lines=left,
    axis line style={draw=black!40, line width=0.4pt},
    tick align=outside, tick style={draw=black!40},
    xmajorgrids, grid style={dashed, color=black!10, line width=0.4pt},
    clip=false,
    legend style={at={(0.5,1.04)}, anchor=south, legend columns=2,
                  font=\scriptsize, draw=none, fill=none,
                  /tikz/every even column/.append style={column sep=10pt}},
    legend image code/.code={\draw[#1, fill=#1] (0cm,0cm) circle (1.7pt);},
]
\fill[hydraBlue, opacity=0.09] (axis cs:80.1,0.55) rectangle (axis cs:80.9,4.55);
\draw[black!40, line width=0.8pt] (axis cs:80.9,4)--(axis cs:84.1,4);
\draw[black!40, line width=0.8pt] (axis cs:80.3,3)--(axis cs:81.9,3);
\draw[black!40, line width=0.8pt] (axis cs:80.1,2)--(axis cs:83.7,2);
\draw[black!40, line width=0.8pt] (axis cs:80.2,1)--(axis cs:81.6,1);
\addplot[only marks, mark=*, mark size=2.8pt, color=hydraTeal]
  coordinates {(84.1,4) (81.9,3) (83.7,2) (81.6,1)};
\addplot[only marks, mark=*, mark size=2.8pt, color=hydraBlue]
  coordinates {(80.9,4) (80.3,3) (80.1,2) (80.2,1)};
\legend{Always-Strong, \textbf{HyDRA}}
\node[anchor=east, font=\tiny\bfseries, color=hydraBlue!80!black] at (axis cs:80.9,4) {80.9\,};
\node[anchor=east, font=\tiny\bfseries, color=hydraBlue!80!black] at (axis cs:80.3,3) {80.3\,};
\node[anchor=east, font=\tiny\bfseries, color=hydraBlue!80!black] at (axis cs:80.1,2) {80.1\,};
\node[anchor=east, font=\tiny\bfseries, color=hydraBlue!80!black] at (axis cs:80.2,1) {80.2\,};
\node[anchor=west, font=\tiny, color=hydraTeal!60!black] at (axis cs:84.1,4) {\,84.1};
\node[anchor=west, font=\tiny, color=hydraTeal!60!black] at (axis cs:81.9,3) {\,81.9};
\node[anchor=west, font=\tiny, color=hydraTeal!60!black] at (axis cs:83.7,2) {\,83.7};
\node[anchor=west, font=\tiny, color=hydraTeal!60!black] at (axis cs:81.6,1) {\,81.6};
\end{axis}
\end{tikzpicture}
\caption{\textbf{Language-invariant routing quality.} Quality retention vs.\ \emph{Oracle Routing} by language group (English $N{=}5{,}007$; European $N{=}2{,}077$; CJK $N{=}494$; Other $N{=}462$). HyDRA (blue) is language-invariant, holding $80.1$--$80.9\%$ across all four groups---a $0.8$-point spread (shaded band)---while staying within $1.4$--$3.6$ points of the Always-Strong upper reference (always routing to the strongest single model, GPT-5.3-Codex, teal).}
\label{fig:lang_invariance}
\end{figure}

\begin{table}[t]
\centering
\small
\begin{tabular}{lccc}
\toprule
\textbf{Language} & \textbf{QR (\%)} & \textbf{$\Delta$QR} & \textbf{CS (\%)} \\
\midrule
English (en) & 80.9 & --- & 37.7 \\
\midrule
Spanish (es) & 80.1 & $-0.7$ & 35.0 \\
Portuguese (pt) & 81.5 & $+0.7$ & 31.5 \\
Chinese (zh) & 79.2 & $-1.7$ & 44.3 \\
French (fr) & 78.8 & $-2.0$ & 33.7 \\
Russian (ru) & 81.2 & $+0.4$ & 43.3 \\
Vietnamese (vi) & 78.2 & $-2.7$ & 36.3 \\
German (de) & 79.3 & $-1.6$ & 38.0 \\
Indonesian (id) & 81.8 & $+0.9$ & 31.6 \\
Turkish (tr) & 78.6 & $-2.2$ & 43.3 \\
Italian (it) & 80.5 & $-0.4$ & 48.3 \\
Japanese (ja) & 79.9 & $-1.0$ & 35.7 \\
Arabic (ar) & 81.5 & $+0.6$ & 45.5 \\
Korean (ko) & 84.6 & $+3.7$ & 28.3 \\
Polish (pl) & 84.0 & $+3.2$ & 44.0 \\
Thai (th) & 80.0 & $-0.8$ & 25.0 \\
\bottomrule
\end{tabular}
\caption{Per-language routing quality and cost savings (\textbf{HyDRA-Multi}, $\tau{=}0.05$). QR uses the same definition as Table~\ref{tab:ml_routing} (vs.\ \emph{Oracle Routing}, Eq.~\ref{eq:qr}); $\Delta$QR is the per-language difference vs.\ English. CS is cost savings vs.\ the costliest in-pool model (Claude-Sonnet-4.6).}
\label{tab:ml_per_lang_accuracy}
\end{table}

\section{Benchmark-Derived Capability Profiles}
\label{app:benchmarks}

This appendix details the benchmark results used to construct model capability profiles for each routing dimension, expanding the two-step computation summarized in \S\ref{sec:profiles}.

\paragraph{Benchmark selection and dimension mapping.}
We anchor profiles to four public coding and tool-use suites chosen to span the four capability dimensions: SWE-Bench Verified~\citep{jimenez2024swebench} and LiveCodeBench~\citep{jain2024livecodebench} (each split into Easy/Medium/Hard difficulty subgroups), BigCodeBench~\citep{zhuo2024bigcodebench}, and $\tau^2$-Bench~\citep{barres2025tau2bench} (Airline and Retail domains). Each benchmark contributes only to the dimensions it can plausibly exercise: the three code suites map to reasoning, code generation, and debugging, while $\tau^2$-Bench maps to reasoning and tool use. A benchmark's weight on a dimension it does not exercise is fixed to zero ($\omega_{b,k}{=}0$), so a model's tool-use score is never inflated by code-only evidence and vice versa.

\paragraph{From raw scores to pool-normalized profiles.}
For each dimension $k$, a model's raw capability is the importance- and judge-weighted average of its per-benchmark/subgroup resolution rates (Step~1, \S\ref{sec:profiles}); harder subgroups receive larger importance weights $\alpha_b$ so that the profile reflects performance where the cheap--strong gap is widest. These raw scores are then affinely mapped into the requirement predictor's empirical score band per dimension (Step~2), pinning the weakest in-pool model to the band floor $\beta^{\text{lo}}_k$ and the strongest to the band ceiling $\beta^{\text{hi}}_k$. This pool-relative normalization is what makes the profiles directly comparable to predicted requirements during shortfall matching, and is also what lets a catalog change be absorbed by re-running only this computation---no retraining of the predictor.

Table~\ref{tab:bench_all} reports per-benchmark scores, Table~\ref{tab:capability_weights} gives the per-benchmark/subgroup and per-dimension weights that combine those scores, and Table~\ref{tab:profiles_summary} shows the resulting pool-normalized profiles used for shortfall matching.

\begin{table*}[!t]
\centering
\small
\begin{tabular}{l ccc ccc r rr}
\toprule
 & \multicolumn{3}{c}{\textbf{SWE-Bench Verified}} & \multicolumn{3}{c}{\textbf{LiveCodeBench}} & \textbf{BigCode} & \multicolumn{2}{c}{\textbf{$\tau^2$-Bench}} \\
\cmidrule(lr){2-4} \cmidrule(lr){5-7} \cmidrule(lr){9-10}
\textbf{Model} & \textbf{Easy} & \textbf{Med.} & \textbf{Hard} & \textbf{Easy} & \textbf{Med.} & \textbf{Hard} & \textbf{Bench} & \textbf{Air.} & \textbf{Ret.} \\
\midrule
Claude Sonnet 4.6 & 85.0 & 71.3 & 44.4 & 100.0 & 86.5 & 38.8 & 52.5 & 64.0 & 81.6 \\
GPT-5.4           & 85.0 & 70.5 & 40.0 & 100.0 & 76.9 & 41.2 & 51.9 & 60.0 & 72.5 \\
GPT-5.3 Codex     & 86.6 & 69.7 & 37.8 & 100.0 & 80.8 & 45.0 & 49.0 & 46.7 & 59.7 \\
GPT-5.2           & 85.0 & 70.5 & 37.8 &  97.7 & 75.0 & 43.8 & 49.1 & 56.6 & 71.5 \\
GPT-5.4 mini      & 80.4 & 67.4 & 33.3 &  97.7 & 67.3 & 30.0 & 49.6 & 45.3 & 62.9 \\
Claude Haiku 4.5  & 80.4 & 67.8 & 31.1 &  93.0 & 46.2 & 15.0 & 49.5 & 60.7 & 72.8 \\
GPT-5             & 80.4 & 63.2 & 31.1 &  97.7 & 50.0 &  7.5 & 42.9 & 46.0 & 58.1 \\
GPT-5-mini        & 75.3 & 51.0 & 17.8 &  97.7 & 69.2 & 26.2 & 42.5 & 46.5 & 58.7 \\
\bottomrule
\end{tabular}
\caption{Per-benchmark resolution rates (\%) used for capability profile derivation. SWE-Bench Verified and LiveCodeBench are split into Easy / Medium / Hard tiers; BigCodeBench and $\tau^2$-Bench (Airline / Retail) report a single aggregate.}
\label{tab:bench_all}
\end{table*}

\begin{table*}[!t]
\centering
\small
\resizebox{\textwidth}{!}{%
\begin{tabular}{l ccc ccc r rr}
\toprule
 & \multicolumn{3}{c}{\textbf{SWE-Bench Verified}} & \multicolumn{3}{c}{\textbf{LiveCodeBench}} & \textbf{BigCode} & \multicolumn{2}{c}{\textbf{$\tau^2$-Bench}} \\
\cmidrule(lr){2-4} \cmidrule(lr){5-7} \cmidrule(lr){9-10}
\textbf{Weight} & \textbf{Easy} & \textbf{Med.} & \textbf{Hard} & \textbf{Easy} & \textbf{Med.} & \textbf{Hard} & \textbf{Bench} & \textbf{Air.} & \textbf{Ret.} \\
\midrule
Benchmark weight $\alpha_b$ & 0.043 & 0.086 & 0.171 & 0.043 & 0.086 & 0.171 & 0.300 & 0.050 & 0.050 \\
\midrule
Reasoning $\omega_{b,1}$ & 0.533 & 0.708 & 0.725 & 0.317 & 0.742 & 0.925 & 0.464 & 0.892 & 0.882 \\
Code Gen $\omega_{b,2}$  & 0.358 & 0.425 & 0.492 & 0.367 & 0.508 & 0.633 & 0.581 & --- & --- \\
Debugging $\omega_{b,3}$ & 0.467 & 0.542 & 0.558 & 0.350 & 0.567 & 0.725 & 0.461 & --- & --- \\
Tool Use $\omega_{b,4}$  & 0.558 & 0.567 & 0.608 & --- & --- & --- & --- & 0.697 & 0.895 \\
\bottomrule
\end{tabular}%
}
\caption{Per-benchmark/subgroup weights used to derive the capability profiles in Table~\ref{tab:profiles_summary} (columns mirror Table~\ref{tab:bench_all}). The raw per-dimension capability of model $m$ refines Step~1 of Algorithm~\ref{alg:capability} as $\text{raw}_{m,c} = \big(\sum_b \alpha_b\, \omega_{b,c}\, s_{m,b}\big) / \sum_b \alpha_b$, where $s_{m,b}$ is the resolution rate (Table~\ref{tab:bench_all}), $\omega_{b,c}$ is the LLM-judge panel's mean dimension weight (``---'' marks a dimension the benchmark does not exercise, i.e.\ $b \notin \mathcal{B}(c)$), and $\alpha_b$ is the benchmark/subgroup importance: parent weights $0.30/0.30/0.30/0.05/0.05$ for SWE-Bench Verified / LiveCodeBench / BigCodeBench / $\tau^2$-Airline / $\tau^2$-Retail, split across Easy/Medium/Hard tiers by difficulty reward $1\!:\!2\!:\!4$ and renormalized so $\sum_b \alpha_b = 1$.}
\label{tab:capability_weights}
\end{table*}

\begin{table}[!ht]
\centering
\small
\begin{tabular}{@{}lcccc@{}}
\toprule
\textbf{Model} & \textbf{Reas.} & \textbf{Code} & \textbf{Debug} & \textbf{Tool} \\
 & $c_1$ & $c_2$ & $c_3$ & $c_4$ \\
\midrule
Claude Sonnet 4.6 & 0.72 & 0.79 & 0.69 & 0.43 \\
GPT-5.4           & 0.65 & 0.71 & 0.63 & 0.28 \\
GPT-5.3 Codex     & 0.62 & 0.77 & 0.67 & 0.10 \\
GPT-5.4 mini      & 0.43 & 0.39 & 0.33 & 0.02 \\
Claude Haiku 4.5  & 0.29 & 0.02 & 0.00 & 0.12 \\
\bottomrule
\end{tabular}
\caption{Final per-model capability profiles $c_{m,k}$ derived from Table~\ref{tab:bench_all} via Algorithm~\ref{alg:capability}. Values shown are the raw pool-normalized per-dimension scores that enter the shortfall computation; the band-compensated dimension weights $\tilde{\mathbf{w}} = (1.25, 0.69, 0.77, 1.29)$ over (\emph{Reas.}, \emph{Code}, \emph{Debug}, \emph{Tool}), derived from band widths via Eq.~\ref{eq:dimcomp}, are applied separately at routing time and are not pre-multiplied into the table values.}
\label{tab:profiles_summary}
\end{table}

\begin{table}[!ht]
\centering
\small
\begin{tabularx}{\linewidth}{@{}lX@{}}
\toprule
\textbf{Benchmark} & \textbf{Dimensions} \\
\midrule
SWE-Bench Verified         & Reasoning, Code Gen, Debugging, Tool Use \\
LiveCodeBench              & Reasoning, Code Gen, Debugging \\
BigCodeBench               & Reasoning, Code Gen, Debugging \\
$\tau^2$-Bench (Airline)   & Reasoning, Tool Use \\
$\tau^2$-Bench (Retail)    & Reasoning, Tool Use \\
\bottomrule
\end{tabularx}
\caption{Benchmark-to-dimension mapping used for capability profile derivation. Each benchmark contributes only to the dimensions it can plausibly exercise.}
\label{tab:benchmark_mapping}
\end{table}

\FloatBarrier

\end{document}